\documentclass[runningheads]{llncs}

\usepackage[table]{xcolor}
\definecolor{lg}{gray}{0.9}
\usepackage{multirow}
\usepackage[T1]{fontenc}
\usepackage{graphicx}
\usepackage{makecell}
\usepackage{xcolor}

\usepackage{algorithm}
\usepackage{algpseudocode}
\usepackage[ruled,vlined,algo2e]{algorithm2e}

\usepackage{tikz}
\usepackage{subcaption}
\usepackage{appendix}
\usepackage{lscape}
\usepackage[figuresleft]{rotating}
\newcommand*{\doi}[1]{\href{http://dx.doi.org/#1}{doi: #1}}

\begin{document}
\raggedbottom
\title{XAI in the context of Predictive Process Monitoring: Too much to Reveal}
\titlerunning{XAI in the context of Predictive Process Monitoring}

\author{Ghada Elkhawaga\inst{1,2}\orcidID{0000-0001-7801-7310} \and
Mervat Abuelkheir\inst{3}\orcidID{0000-0003-0958-7322} \and
Manfred Reichert \inst{1}\orcidID{0000-0003-2536-4153}}

\authorrunning{G. Elkhawaga et al.}

\institute{Institute for Databases and Information Systems, Ulm University, Ulm, Germany \\ \email{\{ghada.el-khawaga,manfred.reichert\}@uni-ulm.de}\\
\and
 Faculty of Computers and Information, Mansoura University, Dakahlia, Egypt
\and
Faculty of Media Engineering and Technology, German University in Cairo, New~Cairo, Egypt\\
\email{ mervat.abuelkheir@guc.edu.eg}}
\maketitle              % typeset the header of the contribution
\begin{abstract}
Predictive Process Monitoring (PPM) has been integrated into process mining tools as a value adding task. PPM provides useful predictions on the further execution of the running business processes. To this end, machine learning-based techniques are widely employed in the context of PPM. In order to gain stakeholders trust and advocacy of PPM predictions, eXplainable Artificial Intelligence (XAI) methods are employed in order to compensate for lack of transparency of most of efficient predictive models. Even when employed under the same settings regarding data, preprocessing techniques, and ML models, explanations generated by multiple XAI methods differ profoundly. A comparison is missing to distinguish XAI characteristics or underlying conditions that are deterministic to an explanation. To address this gap, we provide a framework to enable studying the effect of different PPM-related settings and ML model-related choices on characteristics and expressiveness of resulting explanations. In addition, we compare how different explainability methods characteristics can shape resulting explanations and enable reflecting underlying model reasoning process.  

\end{abstract}

\section{Introduction}
\subsection{Problem Statement}
Predictive process monitoring (PPM) \cite{b1,b2}, as a use case of process mining, supports stakeholders with predictions about the future of a running business process instance. A process instance represents one specific execution instance out of all possible ones enabled by a business process and defined using a business process model. Process mining \cite{PMBook} and PPM both aim at informing stakeholders of how a business process is currently operating or expected to operate in the near future. However, employing black box techniques does not help achieve this purpose. As stakeholders engagement is at the center of process mining tasks, performance and accuracy are not the only aspects which matter while carrying on a PPM prediction task. While depending on ML models in predicting the future of running business process instances, it becomes necessary to persuade business stakeholders of the validity of reasoning mechanisms followed by a predictive model. Justifying predictions to their recipients enable gaining users' trust, engagement and advocacy of PPM employed mechanisms.

EXplainable Artificial Intelligence (XAI) \cite{imlbook} methods and mechanisms \cite{shapL,PDP,DeepLIFT,LRP,Counterfactual,saliency,LIMEPap,ALE} are put in place to provide explanations of predictions generated by a ML model. However, the aforementioned explanations are expected to reflect how a predictive model is influenced by different choices made through PPM workflow. Moreover, PPM tasks employ specific mechanisms to ensure aligning process mining artefacts to ML models requirements. Therefore, while having different XAI methods addressing explainability needs in the context of PPM, studying how explanations of XAI methods differ is an important step to understand the settings suitable for employing these methods, and how to interpret explanations in terms of the underlying influencing factors. On the other hand, as the number of XAI methods increase, and while they address different purposes and different users needs using varying techniques, there is a need to compare different XAI methods outcomes given the same PPM workflow settings.

\subsection{Contributions}
With an attempt to address the need to understand and gain insights into the application of XAI methods into PPM, this paper presents: 
\begin{itemize}
    \item {Comparison of explanations globally and locally, separated and against each other, using different PPM workflow settings using criteria predefined based on underlying data, predictive models and XAI methods characteristics.}
     \item {A study of explanations generated by three different global XAI methods, and two local XAI methods for predictions of two predictive models over process instances from 27 event logs preprocessed with two different preprocessing combinations.}
\end{itemize}

Section \ref{Section 2} provides background information on basic topics needed for understanding this work. In Section \ref{Section 3}, we highlight the basic research questions investigated in this paper. In Sections \ref{5} and \ref{6}, we discuss the settings of the conducted experiments, experimental results and observations. Section \ref{Section 7} highlights lessons learned and conclusions answering the basic research questions. Related work is illustrated in Section \ref{Section 8}. Finally, we conclude the paper in Section \ref{Section 10}.

\section{Backgrounds} \label{Section 2}
This section introduces basic concepts and background knowledge necessary to understand our work. Section \ref{PPMsec} introduces PPM and associated steps to carry on predictions of information relevant to a running business process. Then, we discuss available explainability methods with an in-depth look into those addressing tabular data as these data type is the focus of this work.

\subsection{Predictive Process Monitoring} \label{PPMsec}
Predictive Process Monitoring (PPM) addresses a critical process mining goal resembled in the need to provide decision makers with predictions of the future of a running business process execution instance. This goal can be realised by building models to generate predictions about running process-related information. Examples of these predictive tasks; or what we denote as PPM tasks; are the next activity to be carried out, time-related information (e.g. elapsed time, remaining time till the end), outcome of the process instance, execution cost, or executing resource \cite{PMBook}. Event logs are the input to PPM tasks, and it documents the execution history of a process terms of traces, each of them representing execution data belonging to a single business process instance.
A trace contains mandatory attributes, such as \emph{Case identifier}, \emph{event class}, and \emph{timestamp} \cite{PMBook}. A trace may contain \emph{dynamic attributes} representing information about a single event. Resources fulfilling tasks and documents associated with each event are examples of dynamic attributes. Besides dynamic attributes there are \emph{Static attributes} which have constant values for all events of a given trace.  

\subsubsection{PPM Workflow}
According to the survey results reported in \cite{b1,b2}, a PPM task follows two stages, an offline and another is online. Each stage has steps, to have the two stages with their steps constituting what we call PPM workflow.
\begin{enumerate}
    \item {\textbf{PPM offline Stage.}} An offline stage starts with \emph{constructing a prefix log.} Contemporary PPM approaches that use ML, take as input a prefix log constructed from the input event log. A prefix log is needed to provide a predictive model with incomplete process instances (i.e., a partial trace) for training. Therefore, prefixes can be generated by truncating process instances in an event log up to a predefined number of events. Truncating a process instance can be done up to the first k events of its trace, or up to k events with a gap (g) step separating each two events, where k and g are user-defined. The latter prefixing approach is called \emph{gap-based prefixing}.\\
    
    The following step is to \emph{preprocess prefixes}, in order to be input to a predictive model. Prefix preprocessing sub-steps include bucketing and encoding (cf. Figure \ref{PrePro}). Prefix bucketing means grouping the prefixes according to certain criteria (e.g., number of activities or reaching a certain state in the process execution). The former criteria are defined by the bucketing technique \cite{b2}. Single, state-based, prefix length-based, clustering, and domain knowledge-based are all examples on prefix bucketing techniques. For example in single bucketing,all prefixes generated from the traces of an event log are treated as a single bucket \cite{b1,b2}. Meanwhile in state-based bucketing, different process execution states are determined from the relevant process model,and prefixes are grouped accordingly into buckets. Encoding is the second sub-step of prefixes preprocessing. Prefix encoding means transforming a prefix to a numerical feature vector that serves as input to the predictive model, either for training or for making predictions. Encoding techniques \cite{b1,b2} include static, aggregation, index-based, and last state techniques. Note that static encoding is always used for static attributes encoding, and should be aggregated with one of the aforementioned techniques for encoding dynamic attributes. 

\begin{figure}
\vspace{-3mm}
\includegraphics[width=\textwidth]{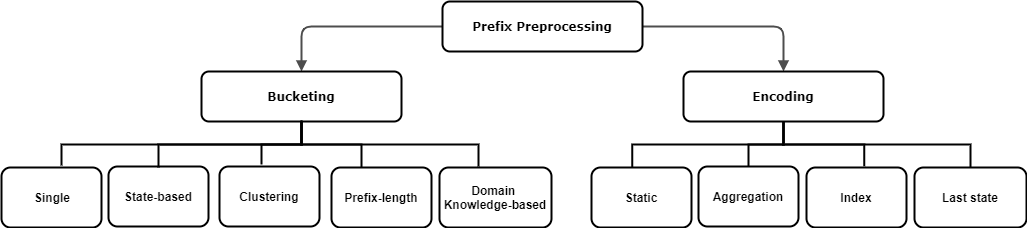}
\caption{Bucketing and Encoding techniques} \label{PrePro}
\vspace{-6mm}
\end{figure}

The following step is to \emph{construct a predictive model.} Depending on the PPM task, the respective predictive model is chosen. The prediction task type, i.e., being classification or regression, is not the only factor guiding a predictive model selection process. Other factors include scalability of the model when facing larger amounts of data, simplicity and interpretability of results. An important step after selecting a model is to find out the suitable set of hyperparameters values, using hyperparameter optimisation. Hyperparameter optimisation \cite{hyperOpt} is the automated process of defining an optimal configuration of hyperparameters that may help increase accuracy. The next sub-step is to train the predictive model on encoded prefixes representing completed process instances. Note that for each bucket a dedicated predictive model needs to be trained, i.e., the number of predictive models depends on the bucketing technique chosen. Finally, after generating predictions for the training dataset, the performance of a predictive model needs to be evaluated.

\item{\textbf{PPM online Stage.}} This stage starts with an incomplete process instance, i.e., a running process instance. Buckets formed in the offline stage are recalled to determine the suitable one for the running process instance. This is accomplished based on the similarity between the running process instance and prefixes in a bucket according to the criteria defined by the bucketing method. Afterwards, the running process instance is encoded according to the encoding method chosen for the PPM task. The encoded form of the running process instance becomes qualified as an input to the prediction method after determining the relevant predictive model from the models created in the offline stage. Finally, the online stage is concluded by the predictive model generating a prediction for the running process instance according to the predefined goal of the PPM task.
\end{enumerate}
\subsection{eXplainable Artificial Intelligence} \label{XAI}
PPM inherits the challenges faced by ML approaches, as a reasonable consequence of employing the latter approaches. One of these challenges concerns the need to gain user trust in the predictions generated. The field of explainable artificial intelligence (XAI) addresses this issue. According to \cite{accountAI}, an explanation is \emph{"a human-interpretable description of the process by which a decision maker took a particular set of inputs and reached a particular conclusion"}. In the context of our research, the decision maker is a ML-based predictive model. Transparency is considered an important aspect of explainability. 
Model transparency can be an inherent characteristic of itself or be achieved through an explanation of a model. Transparent models are understandable on their own and satisfy one or all model transparency levels \cite{XAIResp}. Linear models, decision trees, Bayesian models, rule-based learning, and General Additive Models (GAM) are all considered being transparent/interpretable models. The degree of transparency realised through an explanation may be affected by several factors. These factors include (but are not limited to) the incompleteness of the problem formulation or understanding, the time frame allocated to evaluate the satisfaction of a certain dimension, the level of complexity of an explainability solution, and the number and length of cognitive chunks made available to the user through an explanation.  

\begin{figure}
\includegraphics[width=\textwidth]{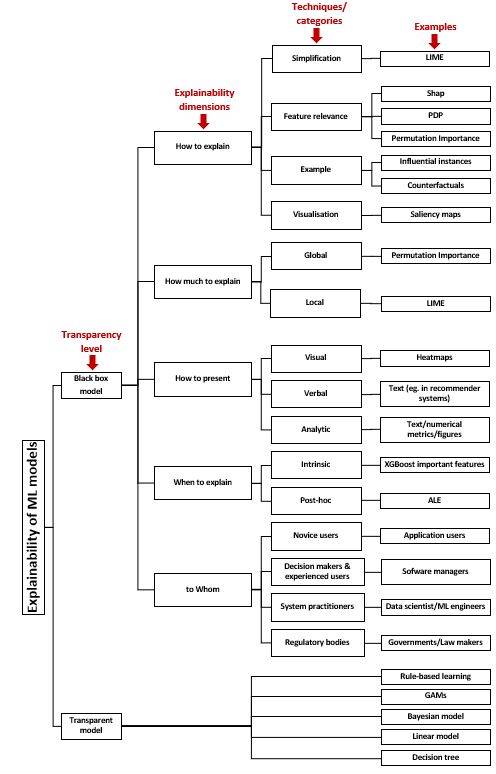}
\caption{Explainability taxonomy in ML} \label{extax}
\end{figure}

Several approaches are proposed under the umbrella of explainability. Explanations construction approaches can be categorised along several dimensions (cf. Figure \ref{extax}). In the following, we illustrate these explainability dimensions. 

\begin{enumerate}
    \item{\textbf{How to explain.}} This dimension is concerned with the approach used to explain how a predictive model derives its predictions based on given inputs. Corresponding approaches have been categorised from different perspectives including design goals and evaluation measures \cite{multidiscip},  transparency of the explained model and explanation scope \cite{XAIResp}, granularity \cite{EvalQual}, and relation to the black box model \cite{survMblack}. For example, a group of approaches tend to generate an \emph{explanation by simplification}. These approaches simplify a complex model by using a more interpretable model called \emph{surrogate} or \emph{proxy model}. The simplified model is supposed to generate understandable predictions achieving an accuracy level comparable to the black box one. Another group of approaches study \emph{feature relevance}. They aim at tracing back the importance of a feature for deriving a prediction. Another family of approaches tend to \emph{explain by example.} Approaches from this category tend to select  representative samples that allow for insights into the model's internal reasoning \cite{XAIResp,EvalQual}. The final category in this dimension tend to explain through \emph{visualisation}, i.e., intermediate representations and layers of a predictive model are visualised with the aim to qualitatively determine what a model has learned \cite{EvalQual}.
    
    Approaches belonging to this XAI dimension are further categorised as being either \emph{model-agnostic} or \emph{model-specific}. Model-agnostic approaches are able to explain any type of ML predictive model, whereas model-specific approaches can only be used on top of specific models. For example, DeepLift \cite{DeepLIFT} and LRP \cite{LRP} provide explanations for neural network-based predictive models. 
    \item{\textbf{How much to explain.}} An explanation may be generated of a various levels of granularity. An explanation effort can be localised to a specific instance or its vicinity, and it can provide global insights into which factors contributed to the decision of a predictive model. \emph{Local explanations} are generated to study interactions between patterns and to better understand how specific input led to a certain output in a given instance, and must not be generalised to provide insights over larger number of instances \cite{PrincPract}. \emph{Global explanations} in turn, provide insights into patterns used by a predictive model over a large number of instances. The scope of an explanation and, subsequently, the chosen technique depend on several factors. One of these factors is the purpose of the explanation, e.g., whether it shall allow debugging the model or gaining trust into its predictions. Target stakeholders constitute another deterministic factor. For example, a ML engineer prefers gaining a holistic overview of the factors driving the reasoning process of a predictive model, whereas an end user is only interested in why a model made a certain prediction for a given instance.  
    \item{\textbf{How to present.}} Choosing the form an explanation is presented is determined by the way the explanation is generated, the characteristics of the end user (e.g., level of expertise), and the scope of the explanation, the purpose of generating an explanation, (e.g., to visualise effects of feature interactions on decisions of the respective predictive model). \cite{multidiscip} introduces three categories of presentation forms. The first one comprises \emph{visual explanation} that uses visual elements like saliency maps \cite{saliency} and charts to describe deterministic factors of a decision in accordance to the chosen explained perspective of a model. \emph{Verbal explanation} provides another way of presenting explanations where natural language is used to describe model reasoning (e.g., in recommender systems). The final form of presentation is \emph{analytic explanation} where a combination of numerical metrics and visualisations are used to reveal model structure or parameters, e.g., using heatmaps and hierarchical decision trees. 
    \item{\textbf{When to explain.}} This dimension of an explainability approach is concerned with the point in time an explanation shall be provided. Agreeing on explainability as being a subjective topic, and depending on the receiver's understanding and needs, we can regard explainability provision from two perspectives. The first one considers explainability as gaining an understanding of decisions of a predictive model and being bounded by model characteristics. Adopting this perspective, explainability is imposed through mechanisms put in place while constructing the model to obtain a white-box predictive model, i.e., \emph{intrinsic explanation}. Using an explanation method to understand the reasoning process of a  model in terms of its outcomes is called \emph{post-hoc explanation} of a predictive model. The second explainability perspective provides an understanding in terms of the whole reasons behind the mapping process between inputs to outputs. Moreover, it provides a holistic view of input characteristics which led to predictive model decisions. Following this perspective implies placing explainability techniques, including pre-modelling, modelling and post-modelling explainability techniques, throughout the whole development pipeline. 
    \item{\textbf{Explain to Whom.}} Studying the target group of each explainability solution becomes necessary to tailor the explanations and to present them in a way that maximizes interpretability of a predictive model and forms a mental model of it. The receivers of an explanation should be at the center of attention when designing an explainability solution. Those receivers can be categorised into different user groups including novice users, decision makers and experienced users, system practitioners, and regulatory bodies \cite{XAIResp,multidiscip}. Targeting each user group with suitable explanations contributes to achieving the explanation process purpose. The purpose of an explanation can be about understanding how a predictive model works or how it makes decisions, or which patterns are formed in the learning process, or which features are crucial to reaching a decision, or any other purpose. Therefore, it is important to understand each user group, identify its relevant needs, and define design goals accordingly.  
    
\end{enumerate}

Note that the various dimensions are tightly interrelated, i.e., a particular choice in one dimension, might affect choices made in other dimensions. Making choices on the different dimensions of the presented taxonomy is guided by several factors. These factors include enhancing understandability and simplicity of the explainability solution. In addition, the availability of software implementations of explainability methods and whether these implementations are model-agnostic or specific, might be deterministic factors of the choice. Moreover, the type of output of each explanation method and hence choosing a suitable presentation type which suits the target group, are also additional factors to be considered. Having explanations serving different user groups with different explainability goals mandates putting explainability evaluation techniques in place to ensure obtaining explanations serving their intended purpose.

\section{Research Questions}\label{Section 3} 
The goal of this research is to study how explanations are affected by underlying PPM workflow-related choices. It is crucial to study different XAI methods characteristics influencing the final outcome of explanation process. Overall, this leads to the following research questions (RQ)s:

\noindent \textbf{RQ1: How can different XAI methods be compared?} 
Conducting a benchmark study that allows comparing all available explainability approaches is not likely to be possible \cite{quanAspect}. This is due to the varying characteristics of these approaches in terms of the dimensions, as mentioned in Section \ref{XAI}. However, as many techniques have been proposed for evaluating explanations \cite{Qinter,LIMESTability}, a constrained study would be useful to compare the relative performance of explainability methods that have been applied in the context of PPM approaches. In addition, the consistency of explainability methods applied to PPM results needs to be studied in order to shed light on one of the potential vulnerabilities of XAI methods, namely sensitivity of explanations. To this end, we subdivide \textbf{RQ1} into the following sub-research questions: \\
\emph{RQ1.1: To what extent are explanations consistent when executing an explanation method several times using the same underlying settings?}\\
\emph{RQ1.2: How are the explanations which are generated by a XAI method affected by predictive model choices?}\\
Explainability methods vary by the extent to which an explanation can be generalised over several data samples that belong to the same vicinity. Explainability methods also vary in the granularity of the explanations they provide and their suitability to explain a number of data samples independent of whether they are small or large, i.e., independent of whether the explainability method is local or global. However, note that the number of explained data samples comes with a computational cost. Therefore, we add another sub-research question to compare explainability methods with respect to their execution time: \\
\emph{RQ1.3: How do explainability methods differ in terms of execution time?}, and how needed time differs with respect to different dataset characteristics, preprocessing choices, and chosen predictive models.

\section{Experiments} \label{5}
This section describes choices and findings of the experiments we performed to compare basic XAI methods. For the basic infrastructure of a PPM outcome prediction task, we are inspired by the framework and findings demonstrated in \cite{b2} and available at \cite{b2git}. Note that we are not changing any of the steps carried out through the aforementioned framework. Settings preservation is needed to be able to observe the impact of the settings described in \cite{b2}, given the reported performance of studied predictive models and preprocessing techniques, from an explainability perspective.

\begin{figure}
%\vspace{-2mm}
\centering
\includegraphics[width=0.7\textwidth, height=6cm]{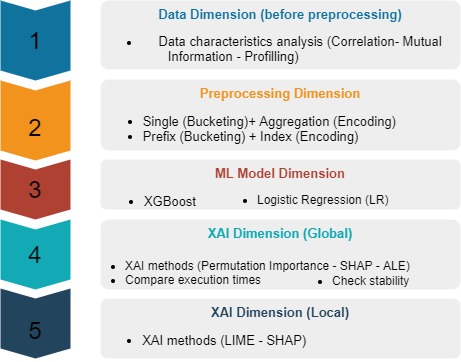}
\caption{Experiments Taxonomy} \label{ExpTax}
\vspace{-5mm}
\end{figure}

Figure \ref{ExpTax} shows a taxonomy of implemented experiments organized under dimensions resembling a ML model creation pipeline, being aligned with the PPM offline workflow, and incorporating an explainability-related dimension. These dimensions are a means to categorise our experiments with the aim of answering the research questions introduced in Section \ref{Section 3}. These dimensions are further discussed through this section. All experiments were run using Python 3.6 and the scikit-learn library \cite{scikitlearn} on a 96 core of a Intel(R) Xeon(R) Platinum 8268 @2.90GHz with 768GB of RAM. Note that we apply all available combinations from each dimension, using the taxonomy defined in Figure \ref{ExpTax} in a dedicated experiment while fixing other options from other dimensions. The code of executed experiments is available through our Github repository\footnotemark[1] \footnotetext[1]{\url{https://github.com/GhadaElkhawaga/PPM\_XAI\_Comparison.git}}to enable open access for interested practitioners. 
  
\subsection{Experimental setup} \label{ExpSetup}
In this subsection, we describe the building blocks of our experiments, including data, chosen preprocessing techniques, selected predictive models, and employed XAI algorithms. Categorising our experiments, and the techniques we employed (as shown in Figure \ref{ExpTax}) is done with the aim of studying the change of explanations. Observing the impact of modifying different parameters at each dimension is planed to the be the basis of this study.
\subsubsection{Data dimension} 
The experiments are carried on three real-life event logs which are publicly available from the 4TU Centre for Research Data \cite{4TU}. The chosen event logs vary in the considered domain (government and banking), the number of traces (representing process instances), and the number of events in each trace. These event logs further vary also in the number of static and dynamic attributes, the number of categorical attributes and, as a result, the number of categorical levels available through each categorical attribute. The three basic event logs \cite{4TU} used are as follows:
\begin{itemize}
    \item {\textbf{Sepsis.}} This event log belongs to the healthcare domain and reports cases of Sepsis as a life threatening condition.
    \item {\textbf{Traffic fines.}} This event log is a governmental one extracted from an Italian information system for managing road traffic fines.
    \item {\textbf{BPIC2017.}} This event log documents load application process in a Dutch financial institution. 
\end{itemize}

\cite{b2} applies several labelling functions to classify each process instance into one of two classes, i.e., a binary classification task. Applying different labelling functions resulted in sub-versions for some of the event logs. These labelling variations result in three extracted logs from the \emph{Sepsis} event log, and three logs extracted from \emph{BPIC2017} event log. These different labelling functions increased the number of used event logs from three to seven event logs. Table \ref{statistics} shows  basic statistics of the event logs used in our experiments. These event logs are cleaned, transformed, and labelled according to the rules defined by the framework available in \cite{b2}.  
\begin{table}
\vspace{-9mm}
\caption{Event logs statistics.}\label{statistics}
\begin{tabular}{|l|l|l|l|l|l|l|l|l|l|l|l|l|l|l|l|}
\hline
 \tiny{\thead{Event\\log}}&\tiny{\thead{\#\\traces}}&\tiny{\thead{Short.\\trace\\len.}}&\tiny{\thead{Avg.\\trace\\len.}}&\tiny{\thead{Long.\\trace}}&\tiny{\thead{Max\\prfx\\len.}}&\tiny{\thead{\#\\trace\\variants}}&\tiny{\thead{\%pos\\class}}&\tiny{\thead{\#\\event\\class}}&\tiny{\thead{\#\\static\\col}}&\tiny{\thead{\#\\dynamic\\cols}}&\tiny{\thead{\#\\cat\\cols}}&\tiny{\thead{\#\\num\\cols}}&\tiny{\thead{\#\\cat\\levels\\static\\cols}}&\tiny{\thead{\#\\cat\\levels\\dynamic\\cols}}\\
\hline
\tiny Sepsis1&\scriptsize 776&\scriptsize 5&\scriptsize 14&\scriptsize 185 &\scriptsize 20&\scriptsize 703&\scriptsize 0.0026&\scriptsize 14&\scriptsize 24&\scriptsize 13&\scriptsize 28&\scriptsize 14&\scriptsize 76&\scriptsize 38\\ \hline
\tiny Sepsis2&\scriptsize 776&\scriptsize 4&\scriptsize 13&\scriptsize 60&\scriptsize 13&\scriptsize 650&\scriptsize 0.14&\scriptsize 14&\scriptsize 24&\scriptsize 13&\scriptsize 28&\scriptsize 14&\scriptsize 76&\scriptsize 39\\ \hline
\tiny Sepsis3&\scriptsize 776&\scriptsize 4&\scriptsize 13&\scriptsize 185&\scriptsize 31&\scriptsize 703&\scriptsize 0.14&\scriptsize 14&\scriptsize 24&\scriptsize 13&\scriptsize 28&\scriptsize 14&\scriptsize 76&\scriptsize 39\\ \hline
\tiny{\makecell{Traffic\\fines}}&\scriptsize 129615&\scriptsize 2&\scriptsize 4&\scriptsize 20&\scriptsize 10&\scriptsize 185&\scriptsize 0.455&\scriptsize 10&\scriptsize 4&\scriptsize 14&\scriptsize 13&\scriptsize 11&\scriptsize 54&\scriptsize 173\\ \hline
\tiny{\makecell{BPIC2017\_\\Accepted}}&\scriptsize 31413&\scriptsize 10&\scriptsize 35&\scriptsize 180&\scriptsize 20&\scriptsize 2087&\scriptsize 0.41&\scriptsize 26&\scriptsize 3&\scriptsize 20&\scriptsize 12&\scriptsize 13&\scriptsize 6&\scriptsize 682\\ \hline
\tiny{\makecell{BPIC2017\_\\Cancelled}}&\scriptsize 31413&\scriptsize 10&\scriptsize 35&\scriptsize 180&\scriptsize 20&\scriptsize 2087&\scriptsize 0.47&\scriptsize 26&\scriptsize 3&\scriptsize 20&\scriptsize 12&\scriptsize 13&\scriptsize 6&\scriptsize 682\\ \hline
\tiny{\makecell{BPIC2017\_\\Refused}}&\scriptsize 31413&\scriptsize 10&\scriptsize 35&\scriptsize 180&\scriptsize 20&\scriptsize 2087&\scriptsize 0.12&\scriptsize 26&\scriptsize 3&\scriptsize 20&\scriptsize 12&\scriptsize 13&\scriptsize 6&\scriptsize 682\\ \hline
\end{tabular}
\vspace{-10mm}
\end{table}

\subsubsection{Preprocessing dimension.} 
Referring back to the preprocessing techniques discussed in Section \ref{PPMsec} and presented in Figure \ref{PrePro}, we make two choices about bucketing and encoding techniques. For bucketing traces of chosen event logs, we apply single bucketing and prefix-length bucketing, with a gap of 5 events. Moreover, we apply aggregation-based and index-based encoding techniques. Normally, both encoding techniques are coupled with static encoding to transform static attributes into a form, in which they can serve as input to a predictive model. As a result, we obtain two combinations of bucketing and encoding techniques namely, \emph{single-aggregation} and \emph{prefix-index}. As index encoding leads to dimensionality explosion due to encoding each categorical level of each feature as a separate column, we decided to apply this encoding on certain event logs, i.e., \emph{Sepsis} (the three derived event logs), \emph{Traffic\_fines} and \emph{BPIC2017\_Refused}. 

\subsubsection{ML model dimension} 
In this study, we employ two predictive models, i.e., XGBoost-based and a Logistic regression (LR). LR can be considered as a transparent model for the same reasons as in the context of a linear regression model, (cf. Section \ref{XAI}). In LR, to each predictor, i.e., feature, a weight is assigned. These weights can be used to indicate how the predictive model has used relevant features in its reasoning process. XGBoost is an ensemble-based boosting algorithm that has proven to be efficient in the context of several PPM tasks \cite{b1,b2}. XGboost is supported by a mechanism to query the model and retrieve a ranked list of important features upon which the model has based its reasoning process.

\subsubsection{XAI dimension}
Referring back to explainability taxonomy from Figure \ref{ExpTax} and the dimensions of an explanation (cf. Section \ref{XAI}), we decided to choose certain explainability methods (\emph{how}), at both explainability levels (\emph{how much}), presented in a certain form (\emph{presentation}), at a certain stage of the predictive model lifetime (\emph{when}). Note that XAI methods we employed in this study fall all under the \emph{model-agnostic} category. Meanwhile, we conducted a complementary study \cite{RQ1to3} conducting experiments using \emph{model-specific} methods. To address the current RQs of this study, we considered only model-agnostic XAI methods. Making choices in this dimension are impacted by certain factors:
\begin{itemize}
    \item The ability of an explainability method to overcome the shortcomings of other methods that explain the same aspects of the reasoning process of a predictive model. For example, Accumulated Local Effects (ALE) \cite{ALE} is adopting the same approach as Partial Dependence Plots (PDP) method \cite{PDP}. However, unlike PDP, ALE takes the effects of certain data characteristics (e.g., correlations) into account when studying features effects \cite{imlbook}.
    \item Comprehensiveness regarding explanation coverage by using local and global explainability methods. Through local explanations, the influence of certain features can be observed. In turn, through global explanations, the reasoning process a predictive model has followed can be inspected. This approach allows reaching conclusions that may provide a holistic view of both, the data and the model applied on the data. Note that it is hard to find a single explainability method that provides explanations at both levels. However, one of the applied methods (i.e., SHAP \cite{shapL}) starts at the local level by calculating contributions of the features on a prediction. It then aggregates these contributions at a global level to give an impression of the impact a feature has on the whole predictions based on a given dataset.
    \item The availability of a reliable implementation of the explainability method. This implementation should enable the integration of the explainability method with the chosen predictive model as well as in the underlying PPM workflow. For example, as we are not using a deep learning model, model-specific explainability methods specialised with deep learning models are not an option in our experiments.  
\end{itemize}
In Table \ref{EXMLMethods}, we categorize each explainability method we applied in the experiments according to the dimensions of an explanation (cf. Section \ref{XAI}). Note that the information in the (\emph{whom}) column which corresponds to the explanations' user groups dimension is case-dependent. Information given in this column is initial and is highly flexible according to several factors including users expertise, application domain and the purpose of the explainability experiment. 
\begin{table} 
\vspace{-8mm}
    \centering
    \caption{Explainability dimensions applied on inspected XAI methods.}\label{EXMLMethods}
  \includegraphics[width=\textwidth, height=3.5cm]{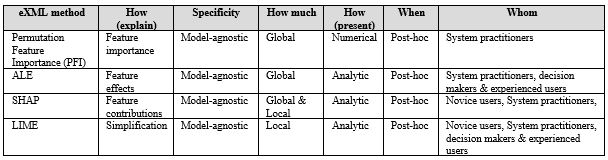}
  \vspace{-15mm}
\end{table}
\subsection{Experiments description} \label{expdesc}
Referring back to the taxonomy from Figure \ref{ExpTax}, we carry out different experiments at each PPM task dimension. For preprocessing and ML model dimensions, experiments conducted are related to analysing the influence of using different techniques (in case of the preprocessing dimension) and the model employed (in case of the ML model dimension). To study the effects of data characteristics, as well as choices made in preprocessing and ML dimensions, we conduct a complementary study concentrating on these factors \cite{RQ1to3}. In this study, we concentrate on how explanations differ based on characteristics and deficiencies of XAI methods. We study how XAI methods highlight the same facts about a predictive model differently. Therefore, here we concentrate on reasoning behind our choices in the XAI dimension. 

\subsubsection{XAI dimension}
Explainability experiments are divided into two sets, global and local experiments, based on the coverage of the explainability method applied, and the type of XAI method used. After executing each group of experiments, we compare the results of the different XAI methods at the level they are applied on. 
\paragraph{Global explainability analysis} 
In this set of experiments, we aim at understanding how a predictive model learns patterns from the training subset of an event log it is fitted to. We use training subsets for two reasons. First, we want to understand how much the model relies on each feature for making predictions. This can be achieved by  studying the change in model accuracy after modifying a certain feature value. To this end, training subsets are more qualified to provide insights into this aspect. Second, we include two model-specific methods in our study, which build their outcomes through the training phase of the predictive model, i.e., based on the training subset of the event log. We have to unify our study of XAI methods outcomes to be based on training subsets of event logs. This decision is made in order to be able to compare the outcomes of model-specific methods with the ones of model-agnostic methods used in the same set of experiments. 

As mentioned previously in this section, Permutation Feature Importance (PFI), ALE and SHAP (the global form) are the model-agnostic methods we used. To check stability of executions, we run the whole experiments taxonomy with different settings twice. Stability checks following this definition are expensive to run, due to expensive computational costs. These costs are affected by the number of datasets with different sizes (where some of the datasets experience dimensionality explosion after the preprocessing phase, and this complicates subsequent explainability steps), and the number of explainability methods applied. As a result, running the whole experiments taxonomy in the context of stability checks was not possible to be done for more than twice. Maybe depending on smaller event logs in the future enables more systematic stability check over higher number of runs. 

To compare the outcomes of all XAI methods we followed the steps illustrated in Algorithm \ref{Algo1}. For PFI and SHAP methods, we study highly correlated features and their importance according to the XAI methods. In addition, we study how the applied model-agnostic XAI methods analyse the importance of features denoted as important to the predictive models through their model-specific explanations, i.e., coefficients in case of Logit and features importance in case of XGBoost. In addition to the aforementioned comparisons, we compare execution times for all applied XAI methods, including time for imitating the explainer and computing features importance as well. We included the training time of the predictive model as the execution time for model-specific methods. 
\paragraph{Local explainability analysis}
In this set of experiments, we apply two model-agnostic XAI methods, namely LIME \cite{LIMEPap} and SHAP \cite{shapL}, analyse their outcomes separately. We apply variable and coefficient stability analysis \cite{LIMESTability} on LIME to study the stability of important feature sets and their coefficients across several runs. 
\section{Results and Observations} \label{6}
It is important to view explanations in the light of all contributing factors, e.g., input characteristics, the effect of preprocessing inputs, and the way how certain predictive model characteristics affect its reasoning process. In \cite{RQ1to3}, we study the effect of different input characteristics, preprocessing choices, and ML models characteristics and sensitivities on resulting explanations. In this section, we illustrate the observations we made during the experiments defined in Section \ref{5}. Due to lack of space, we focus on the most remarkable outputs illustrated by figures and tables. Further results can be generated by running the code of the experiments, which can be accessed via our Github repository. Note that this public availability enables experiments replication and code reusability.
\subsection{Global methods comparability} \label{obsglobexml}
In this subsection, an analysis of PFI, ALE, SHAP results is presented. We execute two runs of each XAI method to query LR and XGBoost models trained over the event logs preprocessed with single aggregation and prefix index combination. Results are compared in order to get insights into stability of results. 
\paragraph{\textbf{Permutation Feature Importance (PFI).}} The basic idea of PFI is to measure the average between the error in prediction after and before permuting the values of a feature \cite{imlbook}. Each of the two PFI execution runs included 10 permutation iterations. The mean importance of each feature is computed. PFI execution led to the following observation:\\
\noindent\fbox{\parbox{\textwidth}{%
In single-aggregated event logs, the results of the two runs are consistent with respect to feature sets and the weights of these features. In prefix-indexed event logs, the two runs are consistent in all event logs.}}\\

\noindent An exception in prefix-indexed event logs is present in ones derived from \\\emph{BPIC2017\_Refused}. In the latter event logs, the dissimilarity between the feature sets across the two runs increases with increasing length of the prefixes. This observation can be attributed to the effect of the increased dimensionality in the event logs with longer prefixes. For prefix-indexed event log, weights of important features change with increasing prefix length.

\paragraph{\textbf{Accumulated Local Effects (ALE).}} ALE \cite{ALE} calculates the change in prediction as a result of changing the values of a feature, while taking features interactions into account \cite{imlbook}. This implies dividing feature values into quantiles \cite{alegithub} and calculating the difference for feature values with a little shift above and below the feature value within a quantile. The described mechanism complicates calculating ALE effects for categorical features, especially one-hot encoded features. However, recently the common python implementation of ALE was modified to compute ALE effects for one-hot encoded features using small values around 0 and 1 \cite{alegithub}. Despite the workaround, computing ALE effects for categorical attributes can be criticised for being inaccurate, because values of these features don't maintain order \cite{imlbook}, \cite{alegithub}.\\ Another issue is that after running ALE over a given event log, effects of each feature are not in a form that yields a rank directly. Therefore, we ranked features based on their entropy. 

\noindent While comparing two execution runs of ALE over the analysed event logs, we make the following observation:\\
\noindent\fbox{\parbox{\textwidth}{%
The encoding technique applied plays a critical role in (dis)similarity between feature ranks in the two execution runs. As an overall observation, ALE tends to be unstable through two runs.}}\\

\noindent In single-aggregated event logs, there is no similarity between the most influencing features in two execution runs. Meanwhile, in prefix-indexed event logs, this observation is not valid in all cases. For example, in prefix-indexed versions of all \emph{Sepsis} logs, there is no similarity between features ranks in both execution runs. Meanwhile, in \emph{Traffic\_fines} and \emph{BPIC\_Refused} encoded using the same combination, the top ranked features are categories derived from the same categorical attributes. Note that due to the inefficiency of ALE to compute effects of categorical attributes, we lean towards a conclusion that a rank where categories derived from categorical attributes dominate, tends to be highly affected by collinearity between categories of such features. This conclusion is valid especially in event logs encoded with a technique which increases the number of categorical attributes exponentially, i.e., index encoding. In aggregation encoding where results from two runs disagree, ranks tend to be more reliable. Aggregation encoding tends to increase the number of ordinal categorical attributes and preserve the existence of numerical attributes. Therefore, feature ranks of a single run tend to be less affected by collinearity. These observations are valid regardless of the underlying predictive model, which enables neutralising the effect of characteristics of the used predictive model.

\paragraph{\textbf{SHapley Additive exPlanations (SHAP).}}
SHAP is an explanation method belonging to the class of feature additive attribution methods \cite{shapL}. These methods use a linear explanation model to compute the contribution of each feature to a change in the prediction outcome with respect to a baseline prediction. Afterwards, a summation of the contributions of all features approximates the prediction of the original model. 
To maintain comparability of the global XAI methods used in our experiments, we constructed a SHAP explainer model on training event logs independently of another SHAP explainer model constructed on relevant testing event logs. The concluded observations made in this section are drawn based on the training SHAP explainer model, whereas the observations based on the testing SHAP explainer model are discussed in Section \ref{compareglobLocal} along with other observations on local XAI methods.\\
\noindent\fbox{\parbox{\textwidth}{%
While comparing the two execution runs, results did not depend on the preprocessing technique used, but differed depending on the predictive model being explained.}}\\

\noindent While explaining predictions of the LR model, performing two executions of the SHAP method did neither result in different feature sets nor different ranks based on SHAP values, regardless the preprocessing combination used. Meanwhile, explaining predictions of XGBoost model reveals having the first most contributing feature as being the same across both execution runs, while the rest of the feature set is the same, but differs in features ranks. An exception is present in the feature set of the three \emph{Sepsis} event logs, where feature ranks are the same across both runs. 
   
\begin{figure}
\vspace{-6mm}
\centering
\includegraphics[width=12cm, height=9cm]{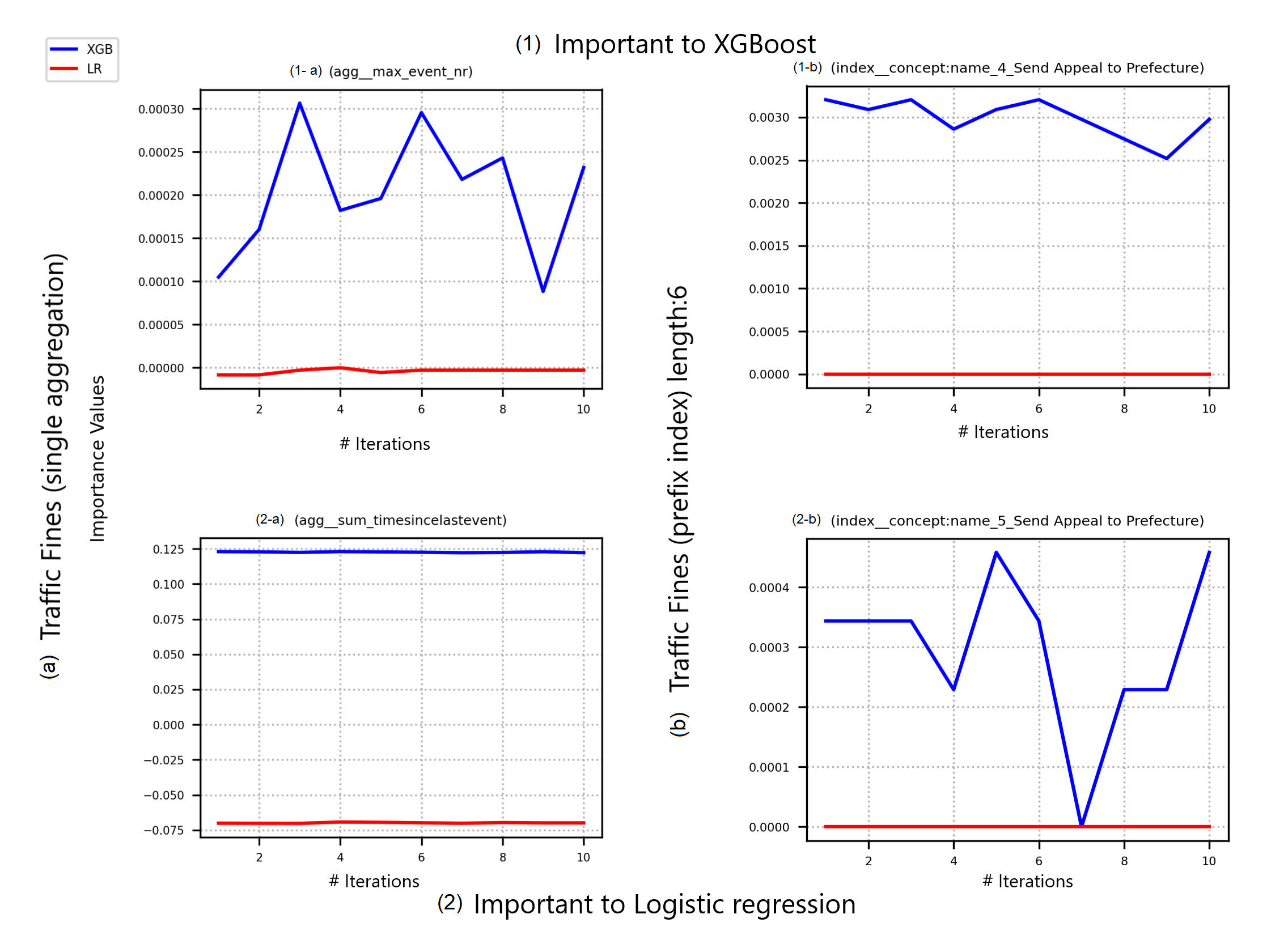}
\caption{PFI scores over \emph{Traffic\_fines} event log} \label{PFItraff}
\vspace{-6mm}
\end{figure}
\subsubsection{Comparability} \label{compareglobLocal}
Studying how PFI, ALE and SHAP analysed the effects or contributions of the most important feature for each predictive model, results in some interesting observations:\\
\noindent\fbox{\parbox{\textwidth}{%
(1) The cardinality of categorical features has an effect on explanations generated by permutation-based methods which depend on measuring prediction error after changing a feature value. The effect of the used preprocessing combination is dependent on the used predictive model.}}\\
  
\noindent For features with lower number of categories, PFI is not able to capture the effect of shuffling feature values on predictions. The effect of shuffling a feature value is static in most event logs predictions using LR, approaching zero in most cases. However, a slight change across PFI iterations can be observed in event logs preprocessed with single aggregation techniques. This observation indicates an effect of the encoding technique along with the interactions between the features. Shuffling values in PFI aims to break dependencies between the features \cite{imlbook}. However, with a low cardinality of a feature, the chances of reducing dependencies decreases over a few number of iterations.

\noindent PFI over XGBoost is presenting slightly higher shuffling effects in prefix-indexed event logs than in LR. This effect is slightly changing over shuffling iterations in the same event logs. However, the increase of changes in XGBoost predictions over iterations of shuffled feature values are observed in prefix-indexed event logs to be higher than in single-aggregated ones. Figure \ref{PFItraff} shows PFI scores for \emph{Traffic\_fines} event log, that was preprocessed with single aggregation (Figures \ref{PFItraff}(1-a), \ref{PFItraff}(2-a)) and prefix index (Figures \ref{PFItraff}(1-b), \ref{PFItraff}(2-b)). Figures \ref{PFItraff}(1-a), \ref{PFItraff}(1-b) represent change in prediction errors for both predictive models while changing the top important features to XGBoost and Figures \ref{PFItraff}(2-a), \ref{PFItraff}(2-b) represent the same for the top two important features to LR. 
    
\begin{figure}
\vspace{-1mm}
\centering
\includegraphics[width=11cm, height=5cm]{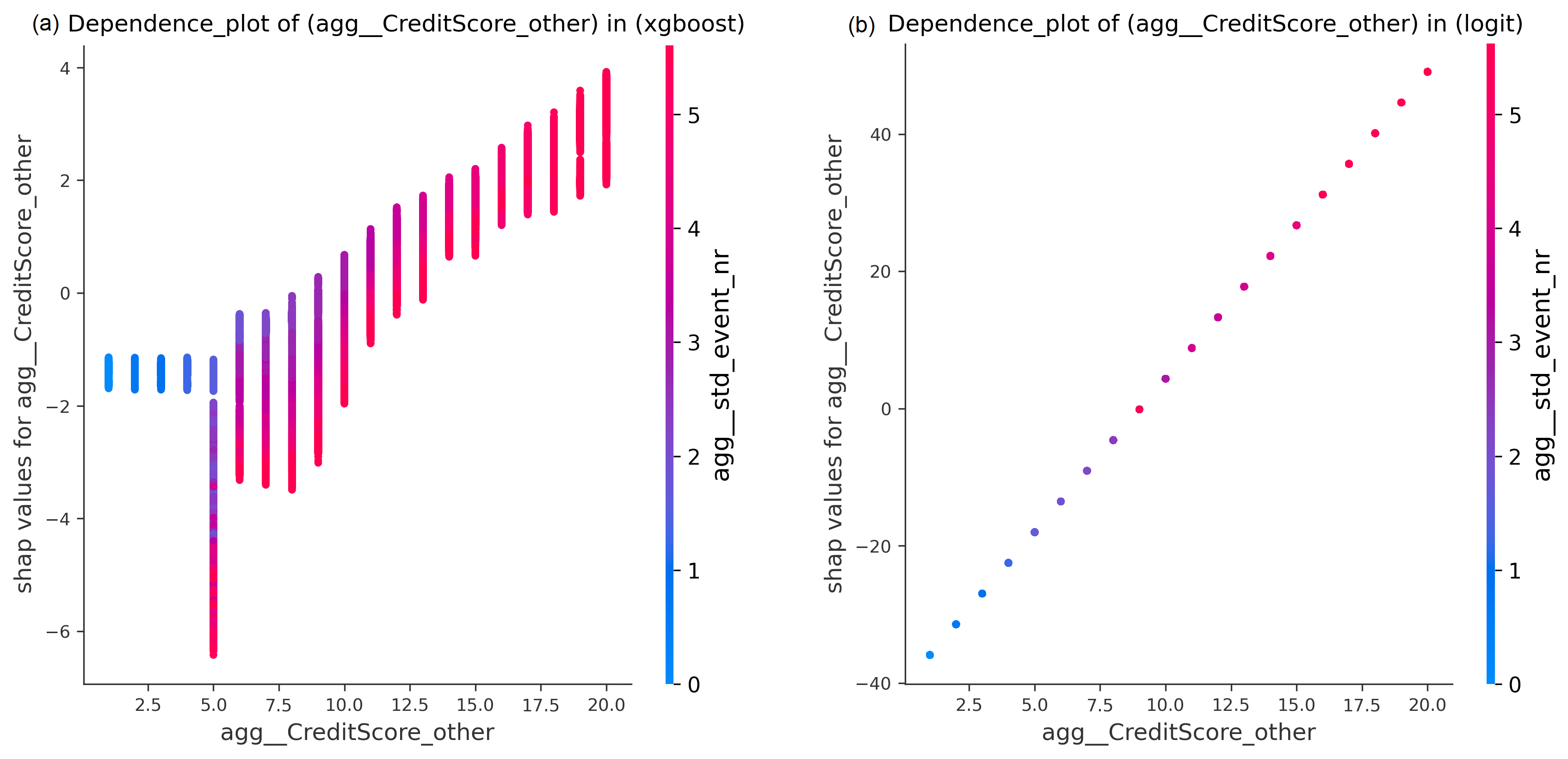}
\caption{SHAP Dependence plots of \emph{BPIC2017\_Refused} (single aggregation) event log} \label{ShapDepBPICRef}
\vspace{-4mm}
\end{figure}

\noindent Dependence plots in SHAP represent an illustrative way presenting the effect of low cardinality in some features. These plots offer analyses of the effect of changing a feature's values on SHAP values while taking the interaction effect of another feature into account. As opposed to PFI, this facility enables acquiring more information from categorical features. For example, Figure \ref{ShapDepBPICRef} shows SHAP values of a feature (CreditScore\_other) according to both, XGBoost and LR predicting outcomes over \emph{BPIC2017\_Refused} event log preprocessed with single aggregation combination. Note that this feature is indicated as the most important one according to LR coefficients and as one of the top five important features based on the XGBoost gain criterion. This feature is a binary feature. However, it has multiple categorical levels due to being encoded based on its frequency of occurrence in a process instance (cf. Section \ref{PPMsec} for encoding types). SHAP values of the feature form a nonlinear curve in XGBoost, and a linear one in LR. In both sub-figures, a point corresponds to a process instance. A point is colored according to its value of an interacting feature, which in this case is "std\_event\_nr". According to LR, shap values increase linearly with increasing "CreditScore\_other" values. The interacting feature values are increasing, as well. However, as a result of having all points of the same "CreditScore\_other" value with identical shap value, it is unclear whether all points of the same "CreditScore\_other" have the same "std\_event\_nr" value. Meanwhile, according to XGBoost, there is an interaction between both features resulting in having points with the same "CreditScore\_other" value to obtain different shap values. These shap values decrease as the values of "std\_event\_nr" increase for points with the same "CreditScore\_other" value.  
     
\noindent\fbox{\parbox{\textwidth}{%
(2) In all event logs, ALE plots for LR are linear themselves.}}\\

\begin{figure}
\vspace{-4mm}
\centering
\includegraphics[width=\textwidth, height=6.5cm]{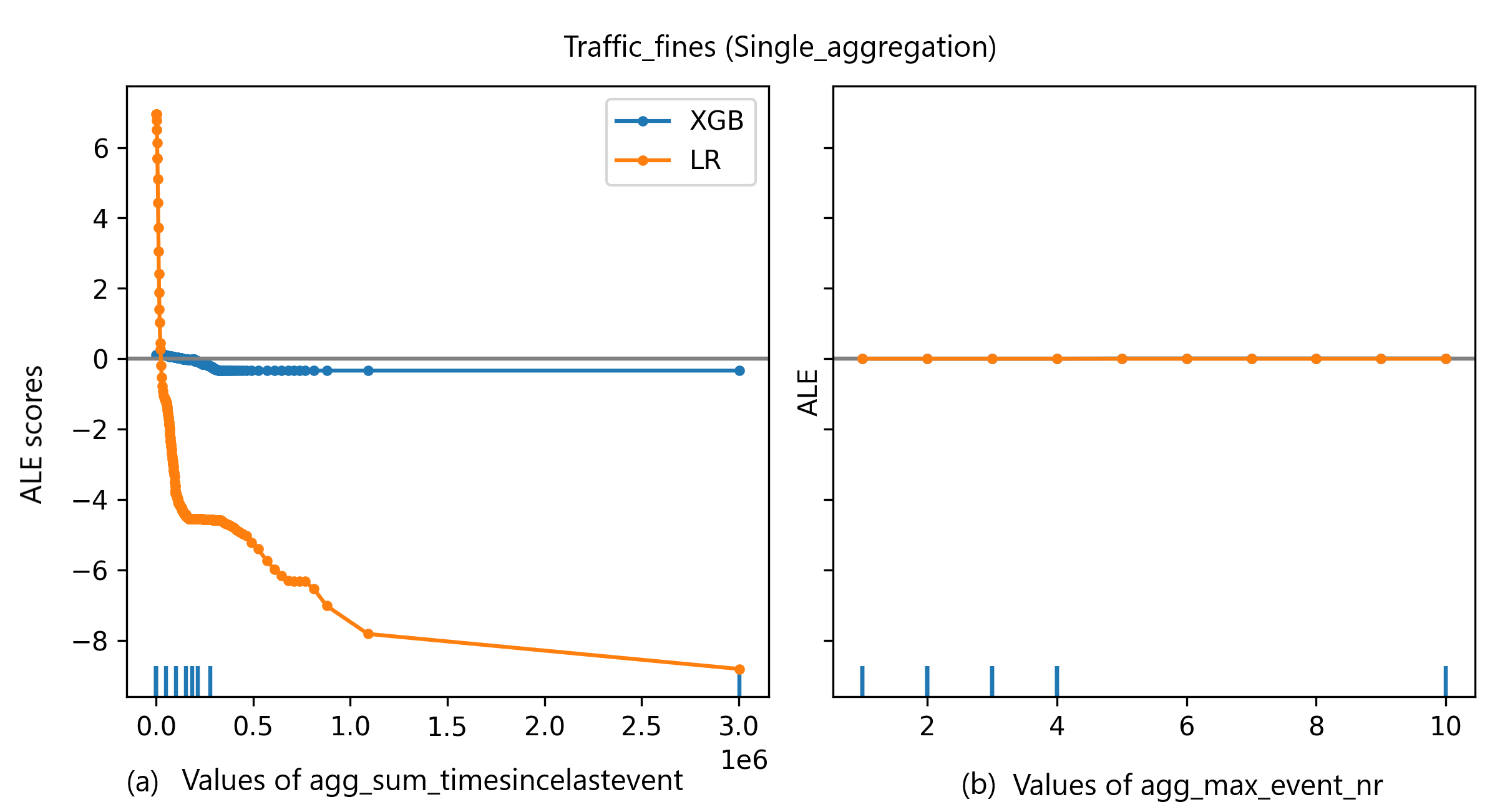}
\caption{ALE plot of \emph{Traffic\_fines} event log} \label{trafficALE}
\vspace{-6mm}
\end{figure}

\noindent An exception to this observation is present in \emph{Traffic\_fines} event logs preprocessed with single aggregation. In figure \ref{trafficALE}(a), the importance of feature "sum\_timesincelastevent", as indicated by LR coefficients, is confirmed by the change in ALE scores with a sudden decrease with increasing feature values. Although Figure \ref{trafficALE} also shows a drecrease in XGBoost predictions, the expected decrease is stable and not steep after "sum\_timesincelastevent" is reaching a value of ($0.5*10^6$) unlike the case with LR model. However, in both cases ALE plot is interpolating as a result of the big gap in feature distribution. Note that the deciles on the x-axis represent interval edges at which ALE scores are calculated. Meanwhile, in between the deciles are parts in which the ALE plot is interpolating. Absence of data
might be an issue in non-linear models, where the ALE plot is interpolating and
hence results may be unreliable in such regions.\\
\noindent\fbox{\parbox{\textwidth}{%
(3) As expected, explaining how a predictive model uses a feature in predicting outcomes of an event log with class imbalance does not reveal a lot of information.}}\\

\begin{figure}
 \vspace{-6mm}
\centering
\includegraphics[width=11cm, height=8cm]{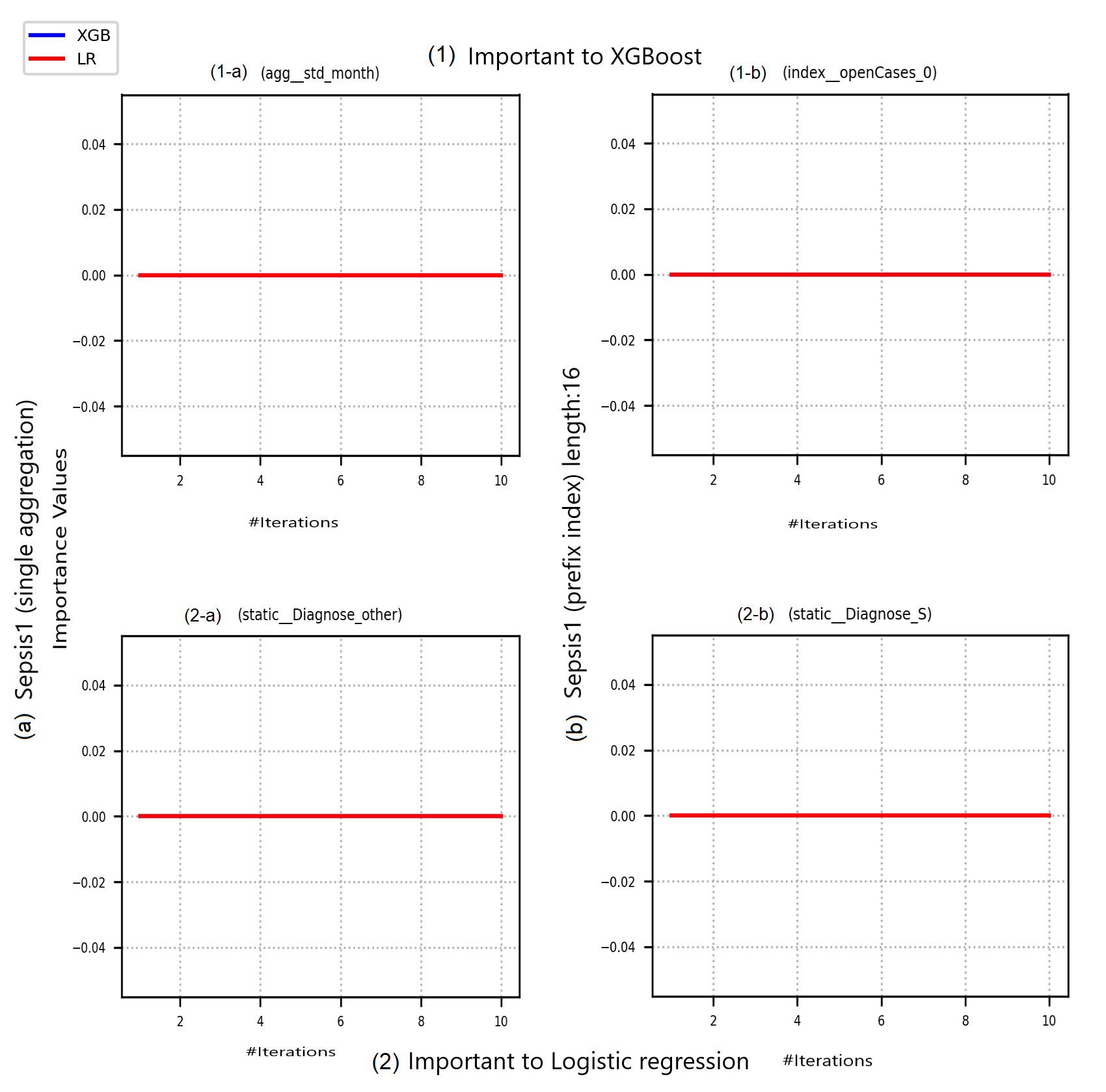}
\caption{PFI scores over \emph{Sepsis1} event log} \label{Sepsis1perm}
\vspace{-3mm}
\end{figure}

\begin{figure}
\centering
\includegraphics[width=12cm, height=9cm]{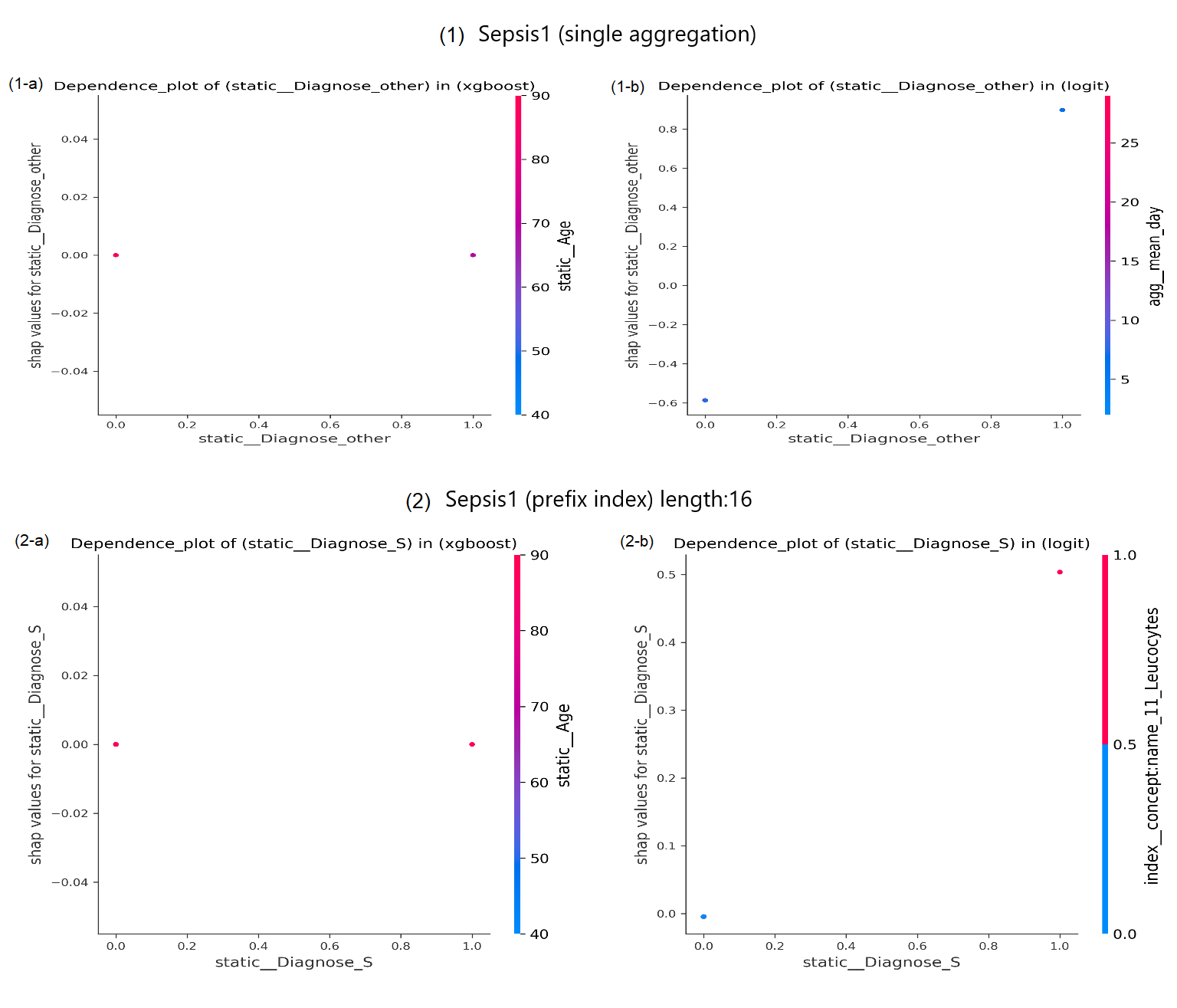}
\caption{SHAP Dependence plots of \emph{Sepsis1} event log} \label{sepsis1shap}
\end{figure}

\begin{figure}
\vspace{-4mm}
\centering
\includegraphics[width=\textwidth, height=6cm]{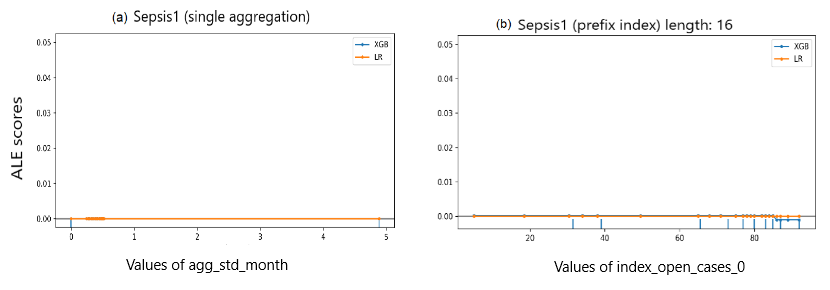}
\caption{ALE scores of \emph{Sepsis1} event log} \label{sepsis1ale}
\vspace{-6mm}
\end{figure}

\noindent This observation is confirmed in results of the three XAI methods, especially when explaining LR reliance on important features. Results of event log \emph{Sepsis1}  shown in Figures [\ref{Sepsis1perm}-\ref{sepsis1ale}] represent an example of this observation.\\
    

\noindent In Figure \ref{Sepsis1perm}, despite the plotted features are indicated as important to the predictive models, after shuffling feature values over 10 iterations, there is no observed effect on the prediction error. Meanwhile, in Figures \ref{sepsis1shap}(1-a) and \ref{sepsis1shap}(2-a), the feature is showing no contribution to change in XGBoost predictions. This observation is represented by the zero SHAP values scored by the categorical features "Diagnose\_other" (Figure \ref{sepsis1shap}(1-b)) and "Diagnose\_S" (Figure \ref{sepsis1shap}(2-b)) in single-aggregated and prefix-indexed versions of \emph{Sepsis1}, respectively.  While a change is observed in LR predictions as a contribution of "Diagnose\_other" in the single aggregation-encoded \emph{Sepsis1}, without effect of the interaction with the feature "mean\_day". However, in Figure \ref{sepsis1shap}(2-b), there is an effect of the interaction between "Diagnose\_S" and\\ "concept:name\_11\_Leucocytes".  It is unclear whether this is the main affecting interaction, as all points with the same value of "Diagnose\_S" provide the same contribution in terms of same SHAP values. ALE as PFI and SHAP are unable to reveal more information about \emph{Sepsis1} and its related event logs. In Figure \ref{sepsis1ale}, both predictive models are not affected by changes in analysed features values, despite being indicated as the most important features. In both sub-figures, the ALE plot is linearly interpolating in between available data. However, ALE scores in both sub-figures are nearly zero.\\
 
\begin{figure}
\vspace{-4mm}
\centering
\includegraphics[width=11cm, height=8cm]{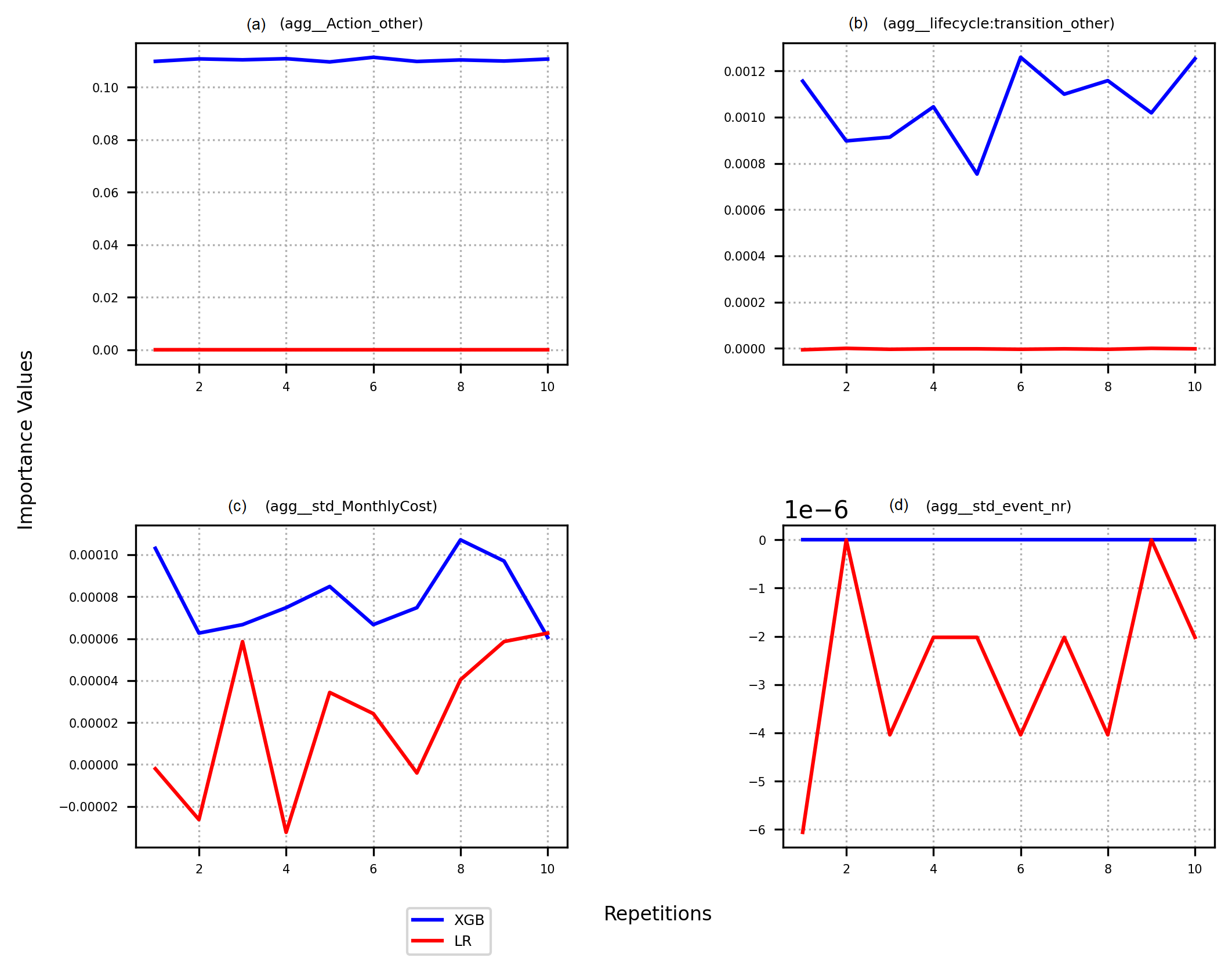}
\caption{PFI scores of \emph{BPIC2017\_Accepted} event log} \label{acceptedperm}
\vspace{-8mm}
\end{figure}

\begin{figure}
\centering
\includegraphics[width=12cm, height=10cm]{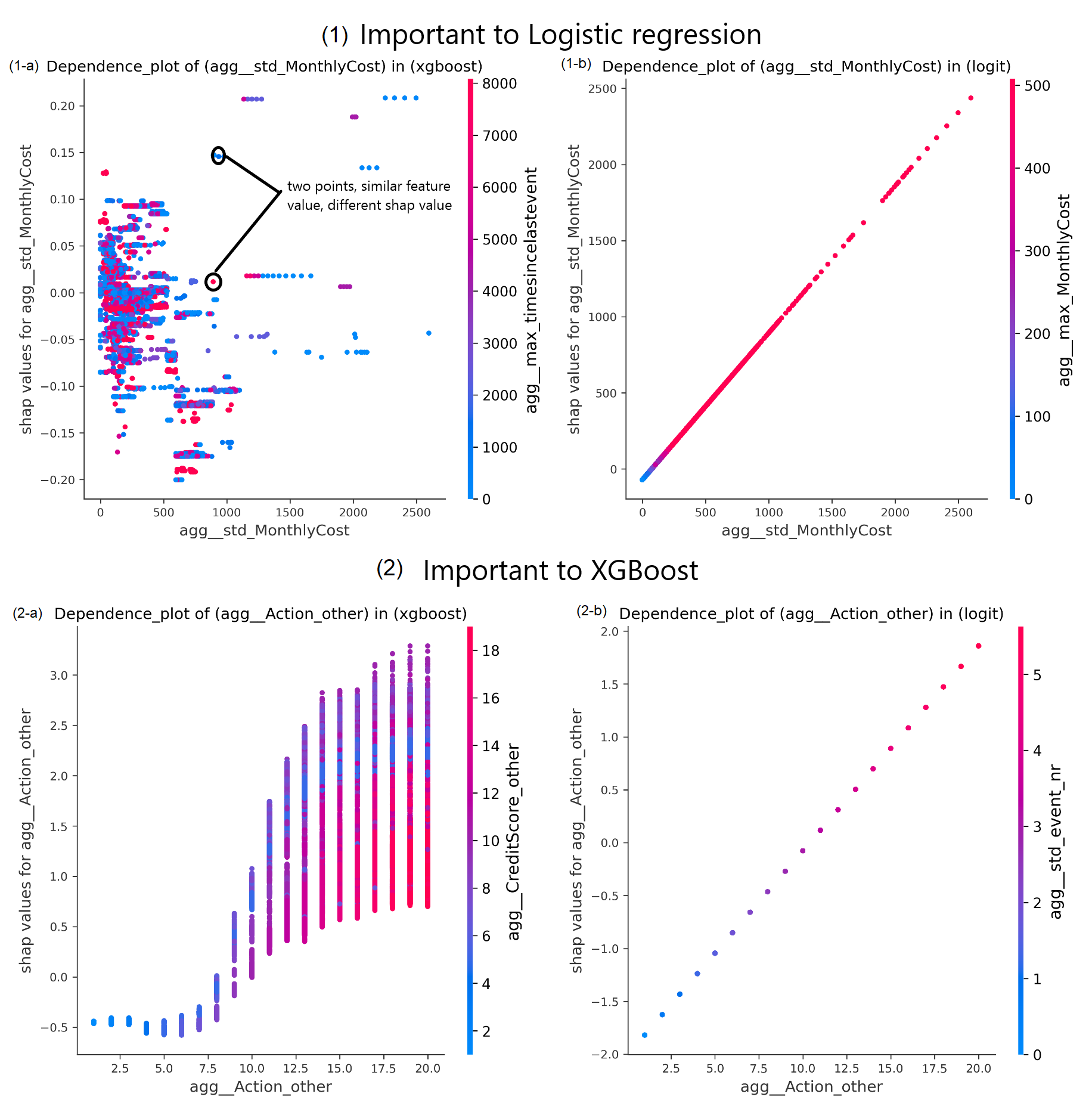}
\caption{SHAP Dependence plots of \emph{BPIC2017\_Accepted} event log} \label{acceptedshap}
\end{figure}

\begin{figure}
\centering
\includegraphics[width=12cm, height=6cm]{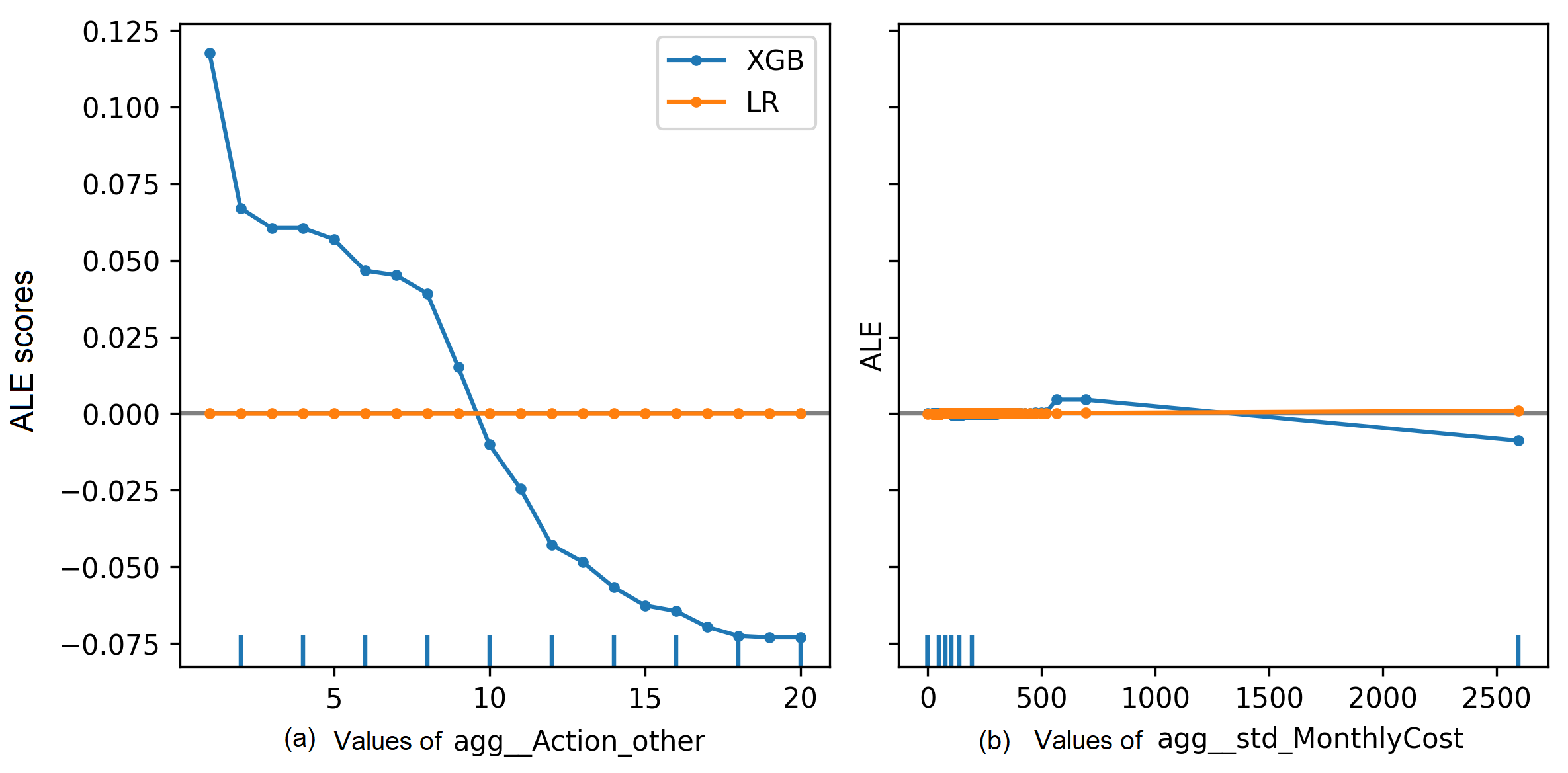}
\caption{ALE scores of \emph{BPIC2017\_Accepted} event log} \label{acceptedale}
\vspace{-5mm}
\end{figure}

As an example, consider the global analysis of \emph{BPIC2017\_Accepted} event log, which is shown in Figures [\ref{acceptedperm}-\ref{acceptedale}]. The high cardinality of the analysed features along with the high number of process instances expose the effect of value change on LR predictions, especially using PFI as shown in Figure \ref{acceptedperm}. The same factors are affecting important features to XGBoost using SHAP and ALE as shown in Figures [\ref{acceptedshap}-\ref{acceptedale}], especially with ALE where the plot is less interpolating with more dispersion of data. In SHAP dependence plots, SHAP values are following a non-linear form, while the interaction effect is clear even in "std\_MonthlyCost" feature which is more important to LR than XGBoost.  


\subsubsection{Execution Times} \label{exectimes}
Execution times for applied XAI methods are computed for SHAP, ALE and PFI over the training event logs. Prediction times of LR and XGBoost are included as the time for computing features important to the predictive models. However, the prediction time is computed as the average of three prediction trials. Execution times are presented in Tables \ref{timesingagg} and \ref{timeprefind}, for execution times of event logs preprocessed with single aggregation and prefix index combination respectively. Note that in the event log column in Table \ref{timeprefind}, each event log is augmented with the length of its prefixes.\\
\begin{table} 
\vspace{-6mm}
\centering
  \caption{Execution times (in seconds) of XAI methods on event logs preprocessed using single aggregation combination.}\label{timesingagg}
  \includegraphics[width=\textwidth, height=3cm]{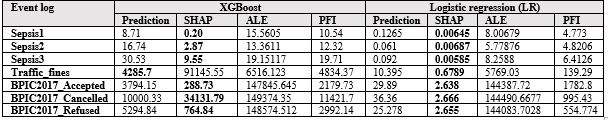}
\vspace{-8mm}
\end{table}
\noindent In terms of the underlying preprocessing technique, the overall numbers across all XAI methods, show that event logs preprocessed with prefix index combination are faster to be explained than those preprocessed with single aggregation.\\
\noindent\fbox{\parbox{\textwidth}{%
(1) As a result, we conclude that the chosen bucketing technique is a determinant factor affecting execution duration of global explanations generation.}} \\

\noindent As illustrated by \cite{b1,b2}, in index encoding, as the prefix length increases, the number of features increases, while the number of process instances decreases in an event log. This fact justifies the faster execution times on prefix length-preprocessed event logs can be based on the reduced number of process instances in an event log compared to a single-aggregated event log. Despite enabling faster executions, as expected, prefix-indexed event logs with longer prefixes show longer execution times over all XAI methods than event logs with shorter prefixes.\\
\noindent\fbox{\parbox{\textwidth}{%
(2) the overall numbers show that explaining decisions of a LR model takes less time compared to an XGBoost-based model.}}\\ 

\noindent This observation is proved to be true by results across different event log sizes, preprocessing settings and over different XAI methods. A justification of this observation may be that all the studied global XAI methods involve querying the underlying model. The boosting mechanism employed in XGBoost complicates the prediction process, and hence demands more time to produce either predictions, or subsequently, explanations.\\
\noindent\fbox{\parbox{\textwidth}{%
(3) While regarding execution times of different XAI methods, it is observed that SHAP has the fastest executions over almost all event logs, regardless the used preprocessing techniques with few exceptions.}}\\

\noindent An exception to this observation is the execution time of SHAP over the shortest prefix version of \emph{Traffic\_fines} event log being preprocessed with prefix index combination and having an XGBoost model trained over its process instances. SHAP values for this event log are the slowest among all XAI methods used over this event log. \\
\noindent\fbox{\parbox{\textwidth}{%
(4) The slowest XAI method differs based on the preprocessing techniques used, while having ALE performing the worst in most cases.}}\\

\noindent In prefix-indexed event logs, PFI achieves the worst performance in most event logs while combined with logit, and ALE is the worst combined with XGBoost. Meanwhile, in single aggregation-based event logs, ALE provides the least efficient XAI method to be combined with either predictive models. PFI is performing the worst in smaller event logs with relevant small numbers of features compared to other logs, e.g., in the three \emph{Sepsis} event logs. 
\begin{table} 
\vspace{-6mm}
\centering
  \caption{Execution times (in seconds) of XAI methods on event logs preprocessed using prefix index combination.}\label{timeprefind}
  \includegraphics[width=\textwidth, height=6cm]{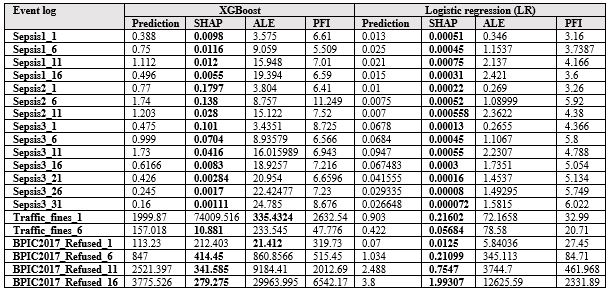}
\vspace{-10mm}
\end{table}
\subsection{Local methods comparability} \label{localcomp}
LIME and SHAP, which are the two local XAI methods used in this study, share the underlying mechanism. Both are feature additive attribution methods \cite{shapL}. LIME is looking for weights of features, to indicate their importance, While SHAP calculates contributions of features to shifting the current process instance prediction towards or apart from a base prediction. To gain insights into how LIME and SHAP results can differ while explaining the same process instances, we compared their explanations of both predictive models reasoning, i.e., LR and XGBoost, over selected set of process instances from each event log preprocessed with both preprocessing combinations. After analysing resulting explanations, the following observations are made:\\
\noindent\fbox{\parbox{\textwidth}{%
(1) It could be rarely observed when both methods match in their attributions to important features or in the strength and direction of the effect features have on driving the current prediction towards or away from the base prediction.}}\\

\begin{figure}
\vspace{-4mm}
\centering
\includegraphics[width=\textwidth, height=8cm]{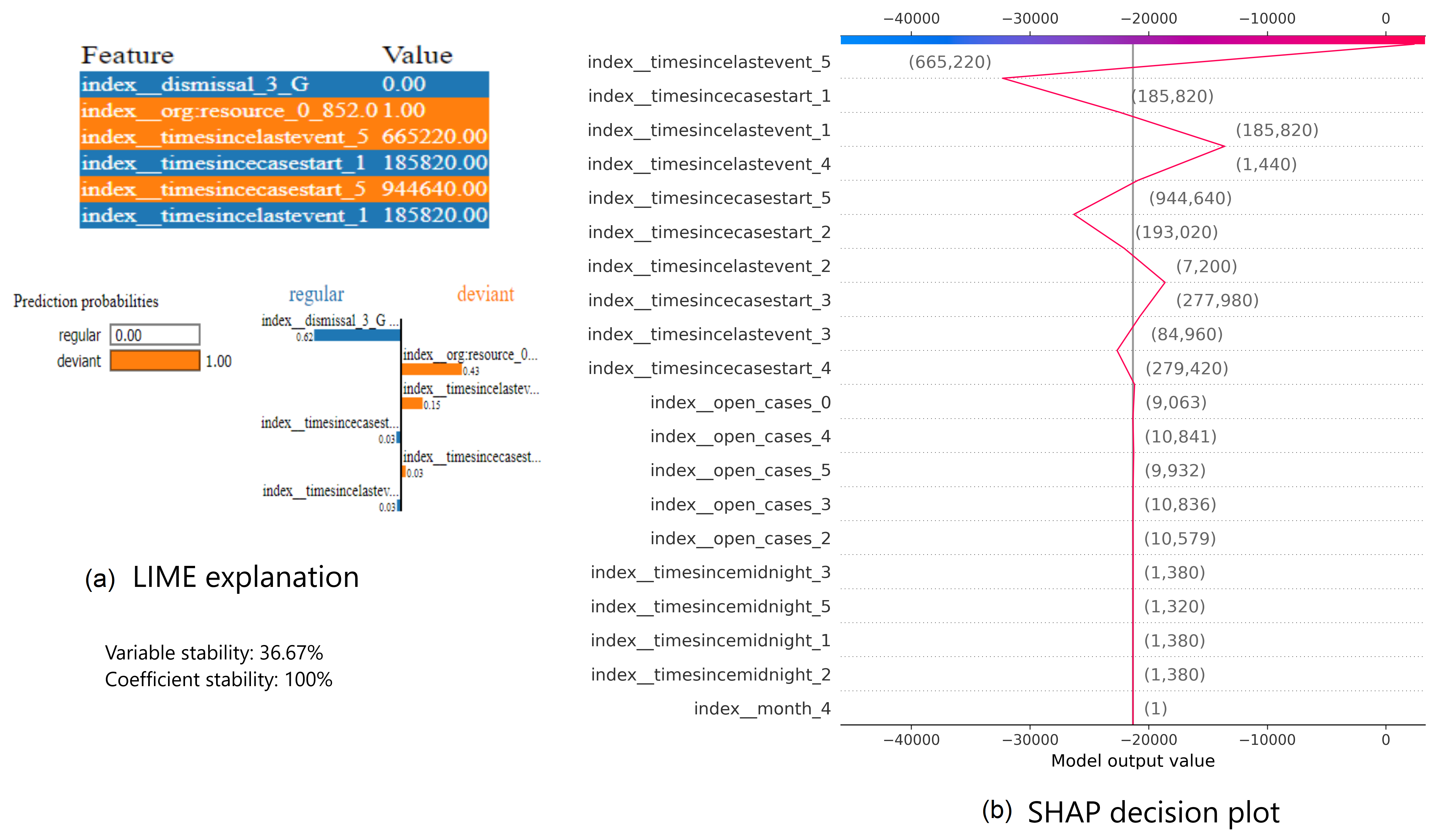}
\caption{LIME vs. SHAP explanations of LR prediction for one instance in \emph{Traffic\_fines} preprocessed with prefix index combination} \label{LIMESHAPlogitprfxidx}
\vspace{-4mm}
\end{figure}

\noindent It is supposed that whenever both XAI methods explain the prediction generated by the same predictive model for the same process instance, a similarity between explanations exists. However, the former observation is valid across all event logs regardless preprocessing combinations used. If both methods agree on a subset of features, they do not have the same importance ranks. Note that in LIME, the goal is to identify a set of features which are important in confirming the current prediction or driving the prediction towards the other class in case of binary classification tasks. Meanwhile in SHAP, the purpose is to order features descending based on their SHAP values, i.e., their contributions in driving the current prediction away or towards the base prediction over the whole event log. 
For example, consider the two explanations in Figure \ref{LIMESHAPlogitprfxidx}. The Figure shows the LIME explanation (Figure \ref{LIMESHAPlogitprfxidx} (a)), and the SHAP decision plot (Figure \ref{LIMESHAPlogitprfxidx} (b)) of prediction made by LR of the same process instance of \emph{Traffic\_fines} event log preprocessed with prefix index. In Figure \ref{LIMESHAPlogitprfxidx} (a), important features are listed in descending order along with their values in the event log  based on coefficients of these features in the approximating model created by LIME. On the bottom of the figure, these features are listed again with the percentage of their contributions. Note that the color code of these features repesent whether a feature is driving the prediction towards the current predicted class or towards the other class. In this case, the currently predicted class is "deviant". 
    
\noindent Figure \ref{LIMESHAPlogitprfxidx}(b) shows the decision plot highlighting the features with the highest shap values in a descending order. The line at the center represents the base prediction or model output. The zigzag line represents the current prediction and it stricks the top at exactly the current prediction value. Predition units are in log odds. The pattern at which the zigzag line is going towards and outwards from the line at the center represents contributions of features at the y-axis to reducing/increasing the difference between the base and current predictions. In addition, feature values are stated in brackets over the zigzag line at the point where the line intersects with the horizontal line. It can be observed that the decision plot is indicating that the current prediction of the LR model is highly affected by collinear features.\\
\noindent\fbox{\parbox{\textwidth}{%
(2) In explanations of both methods, dynamic attributes are dominating feature sets.}}\\

\noindent This observation about explanations can be justified by having all event logs, except for the three dervied from \emph{Sepsis} event log, with a larger number of static attributes and a relatively lower number of dynamic ones (cf. Table \ref{statistics}). Therefore, dynamic attributes as expected are dominating feature sets provided by local explanations. This observation may be justified by the fact that both encoding techniques employed are able to increase the number of features derived from dynamic ones, especially in event logs when the dynamic features dominate. 

\begin{figure}
 \vspace{-7mm}
\centering
\includegraphics[width=\textwidth, height=8cm]{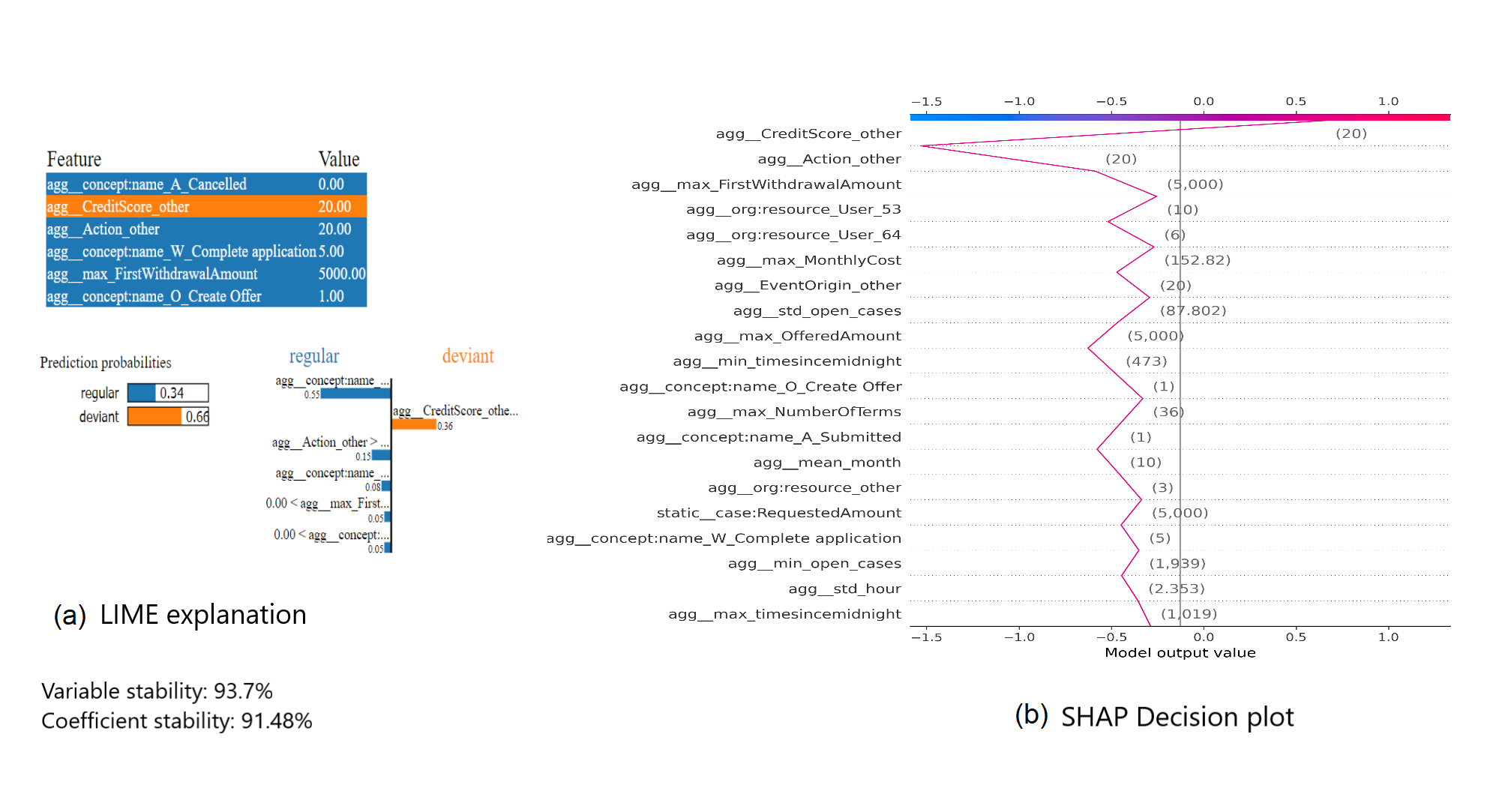}
\caption{LIME vs. SHAP explanations of XGBoost prediction for one instance in \emph{BPIC2017\_Cancelled} preprocessed with single aggregation combination} \label{LIMESHAPXGBoostsingleagg}
\vspace{-5mm}
\end{figure}

\noindent Another example explanation is presented in Figure \ref{LIMESHAPXGBoostsingleagg}. It shows LIME (Figure \ref{LIMESHAPXGBoostsingleagg}(a)) and SHAP (Figure \ref{LIMESHAPXGBoostsingleagg}(b)) explanations of a prediction generated by XGBoost model for a process instance of the \emph{BPIC2017\_Cancelled} event log while being preprocessed with single aggregation combination. In both explanations "CreditScore\_other" is the first or second important feature. In LIME, this feature is the only one driving the prediction towards the currently predicted class. Meanwhile, in SHAP, the same feature has the highest SHAP value and is highest value in driving the current prediction away from the base prediction. This means that this feature has the highest impact according to both explanations.

As mentioned in Section \ref{ExpSetup}, stability measures, namely Variable Stability Index (VSI) and Coefficient Stability Index (CSI) are applied on LIME explanations. VSI \cite{LIMESTability} measures to what extent the same set of important features will be generated if LIME is executed several times over the same event log under the same conditions. Meanwhile, CSI \cite{LIMESTability} measures the stability in coefficients of features within the important feature set over several runs of LIME. \\
\noindent\fbox{\parbox{\textwidth}{%
(3.1) As expected, Variable Stability Index (VSI) for the randomly selected process instances from different event logs show high instability in LIME explanations over several runs. }}\\

\noindent This observation can be justified by the high dimensionality of analysed event logs. According to \cite{LIMESTability}, the underlying approximating model used by LIME, fits a dataset generated based on multivariate distribution of features in the original dataset. The original model is queried for labels of the newly generated dataset. Afterwards, the distance between the newly generated data points and the sample to be explained is measured. In case of high dimensionality, it is not possible to distinguish between distant and near points. The latter phenomena will result in generating a dataset that will differ greatly from the explained sample. As a result, the approximating model is expected to be locally inaccurate compared to the original model \cite{shapL,LIMESTability}, besides having a different approximating model at each LIME run. Consequently, different feature sets will result whenever querying the approximating model. \\
\noindent\fbox{\parbox{\textwidth}{%
(3.2) VSI of LIME explanations of XGBoost predictions are higher in balanced event logs while being higher in imbalanced event logs in case of LIME explanations of LR predictions. }}\\

\noindent For example, VSI for the explanation of the instance in Figure  \ref{LIMESHAPXGBoostsingleagg} is 93.7\%.\\
\noindent\fbox{\parbox{\textwidth}{%
(3.3) CSI measures are high for LIME explanations of XGBoost predictions for all event logs, while the same measures are zeros for explanations of predictions by LR on almost all imbalanced event logs.}}\\

\noindent This observation can be justified by the underlying collinearity of data and sensitivity of linear models; in this case the approximating one; to such phenomena. Consequently, the approximating model is expected to assign unstable coefficients to collinear features at each run of LIME. Therefore, CSI measures are unstable.

\section{Discussion}  \label{Section 7}
Explaining ML-based predictions is a necessity for gaining user acceptance and trust in a predictive models' predictions. It is necessary to regard explainability as a continuous process which should be integrated throughout the whole ML pipeline. A first step towards such integration would be a study of the effect of different pipeline decisions on resulting explanations. Our main concern in this research is to study the ability of an explanation to reflect how a predictive model is affected by different settings in the ML pipeline. Experiments results described and analysed in Section \ref{6} confirm the following conclusions:
\begin{itemize}
    \item Both studied encoding techniques load the event log with a large number of derived features. However, the situation is worse in index-based encoding, as the number of resulting features is increasing proportional to the number of dynamic attributes, especially the number of categorical levels of a dynamic categorical attribute. Dimensionality explosion has an effect on explanations generation as well as predictions generation. On one side, explaining high dimensional event logs becomes expensive in terms of computational resources, especially in XAI methods which run multiple iterations to rank features based on their importance, for example in the case of PFI. We can call this "the horizontal effect of dimensionality". Furthermore, some other XAI methods can not work on lower cardinality features, e.g., ALE , or can work but will not yield useful insights. A high dimensional event log may hold non-useful features which may have been used by the predictive model while can not be used to explain the prediction generated. This can be called "the vertical effect of dimensionality". However, SHAP is the only XAI method among those compared to be able to mitigate the effect of lower cardinality and produce meaningful explanations while highlighting the effect of interactions in dependency plots. These effects are observed in explanations of process instances from event logs preprocessed using index encoding more than aggregation-based preprocessed event logs. 
    
    \item Increased collinearity in the underlying data is another problem resulting from encoding techniques with varying degrees. The effect of collinearity is observed in index-based preprocessed event logs, while not completely absent in aggregation-based event logs. This collinearity is reflected through explanations of predictions on process instances from prefix-indexed event logs as the length of a prefix increases. Another effect of collinearity is the instability of LIME explanations. This instability is due to the approximating model being affected by collinear features (in terms of unstable feature coefficients) and high dimensionality (in terms of unstable feature sets). 
    
    \item PFI showed to be more stable and consistent along two execution runs, while SHAP stability is affected by the underlying predictive model while may be (in)sensitive to underlying data characteristics. ALE is mostly unstable and affected by its inability to accurately analyse effects of changes in categorical attributes on predictions generated.
\end{itemize}
Observations made in the context of this study raise potential opportunities for further research. Setting criteria for evaluating XAI methods is a demanding need to ensure acceptance and trustworthiness of XAI methods themselves. Measurements should be put in place to ensure stability and replicability of results across several runs of a XAI method. Sensitivity of a XAI method to changes in underlying data or underlying predictive model should be measured and an XAI method should be evaluated for. This issue highlights a need to regard an explainability method as an optimisation problem where different underlying choices affect the final outcome, as we prove by results of this study. 

\section{Related Work} \label{Section 8}
Some work available from related research areas is considered complementary to the work we present in this paper. These research efforts may enable gaining solid understanding of how different aspects of predictions explainability are studied in the context of PPM, besides shedding light on possible research gaps.

\subsection{Leveraging PPM with explanations}\label{LeveragePPM}
\cite{XNAP} proposes an approach which integrates Layer-wise Relevance Propagation(LRP) to explain next activity predicted using an LSTM predictive model. This approach tends to propagate relevance scores backwards through the model to indicate which previous activities were crucial to obtaining the resulting prediction.  This approach provides explanations for single predictions, i.e, local explanations. 

Another approach explaining an LSTM decisions is present in \cite{SHapItalian}. However, \cite{SHapItalian} claim that their approach is model-agnostic and not dependent on the chosen predictive model. This approach predicts the execution of certain activities, besides predicting remaining time and cost of a running process instance. The total number of process instances where a certain feature is contributing either positively or negatively to a prediction is identified at each timestamp for the whole dataset. This identification is directed by SHAP values. \cite{SHapItalian} used the same approach in providing local explanations for running process instances. 

Explanations can also be used to leverage a predictive model performance as proposed by \cite{featItems}. Using LIME as a post-hoc explanation technique to explain predictions generated using Random Forest, \cite{featItems} identified feature sets which contributed the producing wrong predictions. After identifying these feature sets, their values are randomised, provided that they don't contribute to generating right predictions for other process instances. The resulting randomised dataset is then used to retrain the model again till its perceived accuracy is improved.

\subsection{Using transparent models in PPM tasks}\label{TransPPM}
Proposals in \cite{attentionNN,flowanalysis} represent attempts to provide a transparent PPM approach, supported by explanation techniques with the aim of providing a transparent predictive model. \cite{attentionNN} conducts experiments on three different predictive models to predict next activity only using control-flow information, next activity supported with dynamic attributes, and next activity along with remaining time. \cite{attentionNN} uses LSTM with attention mechanisms to provide attention values indicating attention weights used to direct the attention of the predictive model to certain features. According to claims in \cite{attentionArg} and \cite{attentionArg2}, there is a lot of argument in the field of Natural Language processing (NLP) about whether to use attention weights as explanations, as attention weights are not always correlated to feature importance \cite{attentionArg}. 

\cite{flowanalysis} proposed a process-aware PPM approach to predict the remaining time of a running process instance, while providing a transparent model. However, \cite{flowanalysis} could not prove that the proposed approach provides comprehensible explanations from the user perspective. Meanwhile, the proposed approach includes discovering a process model, training a set of classifiers to provide probabilities of gates in the discovered process model, and training a set of regressors to perform the main prediction task. These procedures are source of computation overhead which is not confirmed or disproved by the authors of \cite{flowanalysis}. 

\section{Conclusions} \label{Section 10}
In this research, a framework is implemented for comparing explanations generated by a selected number of currently available XAI methods. The XAI methods under analysis explain decisions of ML models generating predictions in the context of PPM. Our study includes comparing XAI methods at different granularity levels, given different underlying PPM workflow decisions. A study of how an XAI method is able to reflect a predictive model's sensitivity towards underlying data characteristics is provided. Meanwhile, different XAI methods are compared according to different criteria including stability and execution duration, over different granularities, i.e., globally and locally. This study has revealed how explanations can highlight data problems through analysing the model reasoning process. It has also emphasized based on experiments, the importance of feature selection after preprocessing an event log. Our study has highlighted situations where data problems may not affect the accuracy of predictions, but do affect usefulness and meaningfulness of explanations. Explainability should be seamlessly integrated into PPM workflow stages as an inherent task not as a follow up effort.

\appendix 
\section{Appendix}
\begin{figure} 
    \centering
  \includegraphics[width=\textwidth, height=\textheight]{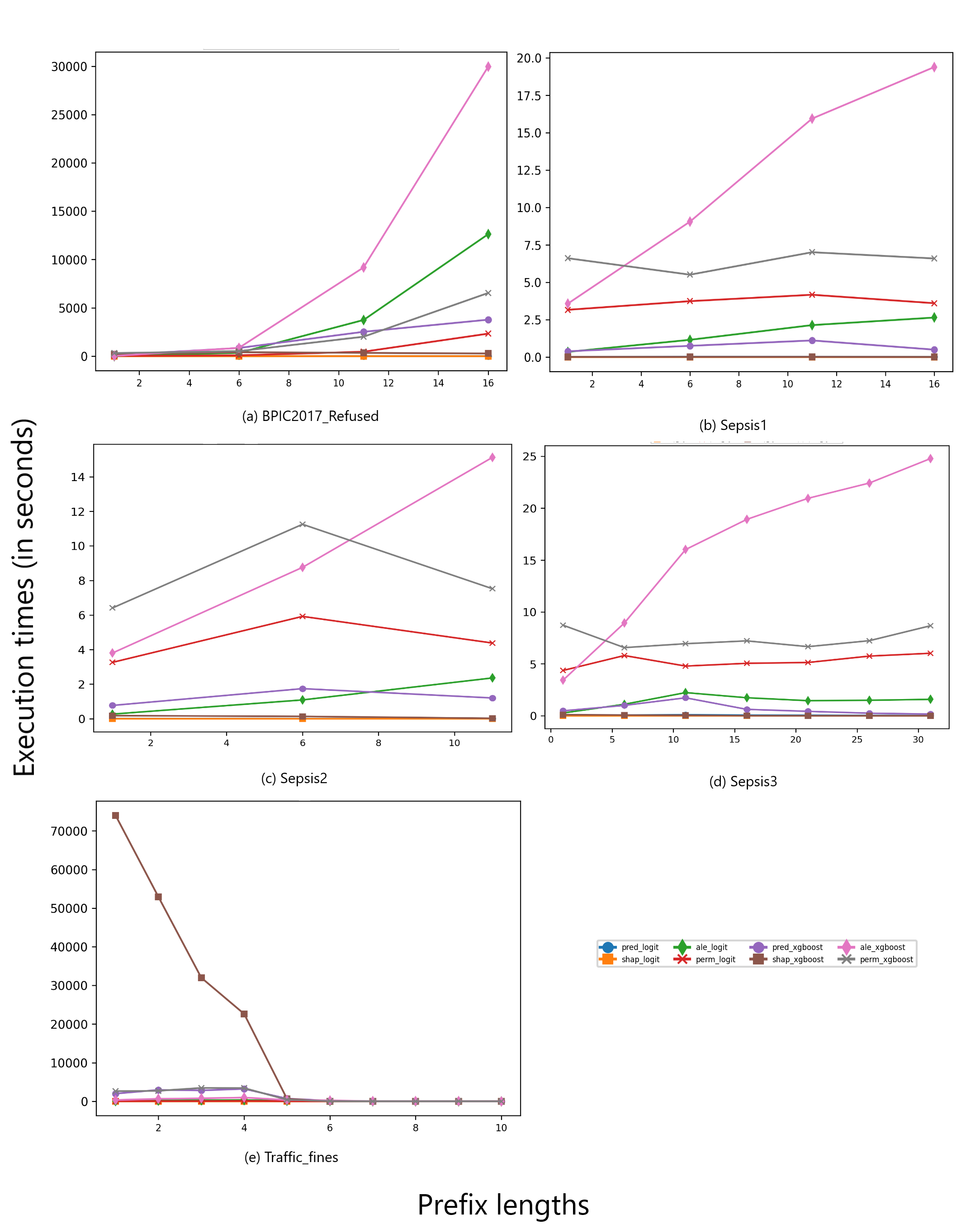}
\caption{Execution times (in seconds) of XAI methods on event logs preprocessed using prefix index combination.}\label{execprfxindxall}
\end{figure}

\begin{figure}
    \centering
  \includegraphics[width=\textwidth, height=\textheight]{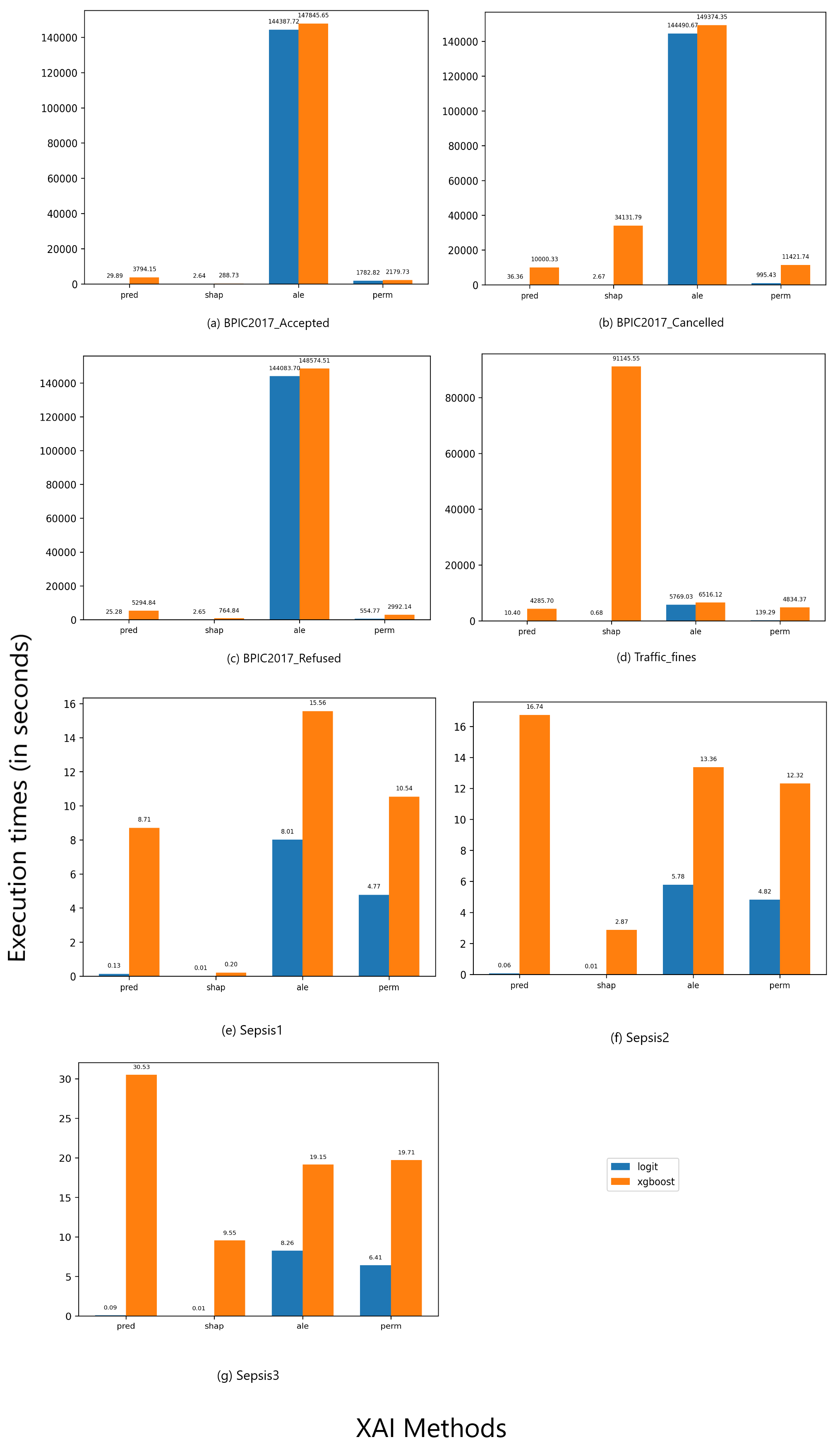}
\caption{Execution times (in seconds) of XAI methods on event logs preprocessed using single aggregation combination.}\label{execsingleaggall}
\end{figure}

This appendix reports the following:
\begin{itemize}
\item Execution times comparison over prefix-indexed event logs with respect to different prefix lengths(Figure \ref{execprfxindxall}).
\item Comparison of Execution times of XAI methods over single-aggregated event logs classified based on predictive models(Figure \ref{execsingleaggall}).
\end{itemize}

\end{document}